\documentclass[]{beingbeyond}
\usepackage{enumitem}
\usepackage[toc,page,header]{appendix}

\usepackage[utf8]{inputenc} 
\usepackage[T1]{fontenc}    
\usepackage{hyperref}       
\usepackage{url}            
\usepackage{array}          
\usepackage{booktabs}       
\usepackage{amsfonts}       
\usepackage{nicefrac}       
\usepackage{microtype}      
\usepackage{xcolor}         
\usepackage{xspace}
\usepackage{bm}
\usepackage{bbm}
\usepackage{bbding}
\usepackage{tabularx}
\usepackage{textcomp}
\usepackage{amssymb}
\usepackage{enumitem}
\usepackage{amsmath}
\usepackage{mathtools}
\usepackage{amsthm}
\usepackage{multirow}
\usepackage{makecell}
\usepackage{color}
\usepackage{colortbl}
\usepackage{adjustbox}
\usepackage{caption}
\usepackage{graphicx}
\usepackage{wrapfig}
\usepackage{array}
\usepackage{multicol}
\usepackage{algorithm}
\usepackage{algorithmic}
\usepackage{diagbox}
\newcommand{\eg}{\textit{e.g.}}

\definecolor{myyellow}{RGB}{255,192,0}
\definecolor{mygreen}{RGB}{107,170,64}
\definecolor{mywrite}{RGB}{255,227,132}

\title{UniTacHand: Unified Spatio-Tactile Representation for Human to Robotic Hand Skill Transfer}

\author{{\bfseries 
Chi Zhang$^{1,2,*}$ \quad 
Penglin Cai$^{1,2,*}$ \quad
Haoqi Yuan$^{1,2}$ \\
Chaoyi Xu$^{2}$ \quad
Zongqing Lu$^{1,2,\dagger}$
}}

\affiliation{{$^{1}$Peking University \quad $^{2}$BeingBeyond}}

\webpage{\url{https://beingbeyond.github.io/UniTacHand/}}

\abstract{
Tactile sensing is crucial for robotic hands to achieve human-level dexterous manipulation, especially in scenarios with visual occlusion. However, its application is often hindered by the difficulty of collecting large-scale real-world robotic tactile data.  
In this study, we propose to collect low-cost human manipulation data using haptic gloves for tactile-based robotic policy learning. The misalignment between human and robotic tactile data makes it challenging to transfer policies learned from human data to robots.
To bridge this gap, we propose \textbf{UniTacHand}, a unified representation to align robotic tactile information captured by dexterous hands with human hand touch obtained from gloves. 
First, we project tactile signals from both human hands and robotic hands onto a morphologically consistent 2D surface space of the MANO hand model. This unification standardizes the heterogeneous data structures and inherently embeds the tactile signals with spatial context. 
Then, we introduce a contrastive learning method to align them into a unified latent space, trained on only 10 minutes of paired data from our data collection system. 
Our approach enables zero-shot tactile-based policy transfer from humans to a real robot, generalizing to objects unseen in the pre-training data. 
We also demonstrate that co-training on mixed data, including both human and robotic demonstrations via \textbf{UniTacHand}, yields better performance and data efficiency compared with using only robotic data.
\textbf{UniTacHand} paves a path toward general, scalable, and data-efficient learning for tactile-based dexterous hands.
}

\checkdata[Date]{December 17, 2025}

\definecolor{BlockC}{gray}{0.98}  
\definecolor{BlockA}{RGB}{191,211,230}
\definecolor{BlockB}{RGB}{199,233,192}

\begin{document}

\maketitle
\begingroup
\renewcommand\thefootnote{\fnsymbol{footnote}} 
\setcounter{footnote}{0}
\footnotetext[1]{Equal contribution.}
\footnotetext[2]{Correspondence to Zongqing Lu $<$lu@beingbeyond.com$>$.}
\endgroup

\section{Introduction}

\label{sec:intro} Dexterous manipulation for contact-rich tasks remains a grand challenge in robotics~\citep{suh2025dexterous, wu2021learning, liang2025dexhanddiff, lakshmipathy2024contactmpc}, largely because the systems lack the rich, multi-modal tactile perception that humans use. While vision-based policies~\citep{diffusion-policy, dexgraspvla, demograsp} are dominant, they struggle with fundamental limitations like occlusion and ambiguity in complex interactions~\citep{levine2018learning, yuan2017gelsight}. Tactile sensing provides a distinct, complementary information stream that addresses these shortcomings~\citep{dahiya2010tactile}. However, despite the advantage of diverse, high-resolution tactile sensors, effectively processing and unifying heterogeneous tactile data remains an open problem~\citep{gao2021survey}.


To endow robots with autonomous manipulation capabilities, data-driven methods have become increasingly prominent. Reinforcement learning (RL) for robotics, while showing promise for some tasks, often suffers from high sample complexity~\citep{kober2013reinforcement} and the well-known sim-to-real gap -- especially pronounced in tasks with complex contact dynamics~\citep{argall2009survey, chebotar2019closing, resdex}.
Imitation learning (IL) and learning from demonstration (LfD) methods have emerged as a promising alternative, directly mimicking expert data to acquire manipulation skills. However, applying IL to tactile-based dexterous manipulation is fundamentally limited by the general scarcity and low quality of real robot teleoperation data~\citep{aloha,doglove}. LfD methods~\citep{hoque2025egodex, chen2022dextransfer, beingh0}, especially those that utilize human data collected by wearable devices, are hindered by the embodiment gap between human and robotic hands~\citep{wang2023cross}.

Current efforts to bridge the embodiment gap have largely focused on kinematic retargeting~\citep{crossdex, hoque2025egodex}, leaving the tactile morphological gap between different robotic hands largely unaddressed. A common workaround is to equip human hands with the same robotic tactile sensors used on the robot for data collection~\citep{wang2023dexumi, FTF}. However, such approaches rely on hardware uniformity, which can be difficult to achieve for many robotic hands and tends to be less cost-effective or scalable for data collection.
In this research, we aim to use human data collected by wearable haptic gloves to learn tactile-based robotic manipulation policies. Our focus thus shifts to the following question: how can we design a method to transfer tactile manipulation knowledge from haptic gloves to dexterous robotic hands?


\begin{figure}[t]
  \centering
  \includegraphics[width=1.0\textwidth]{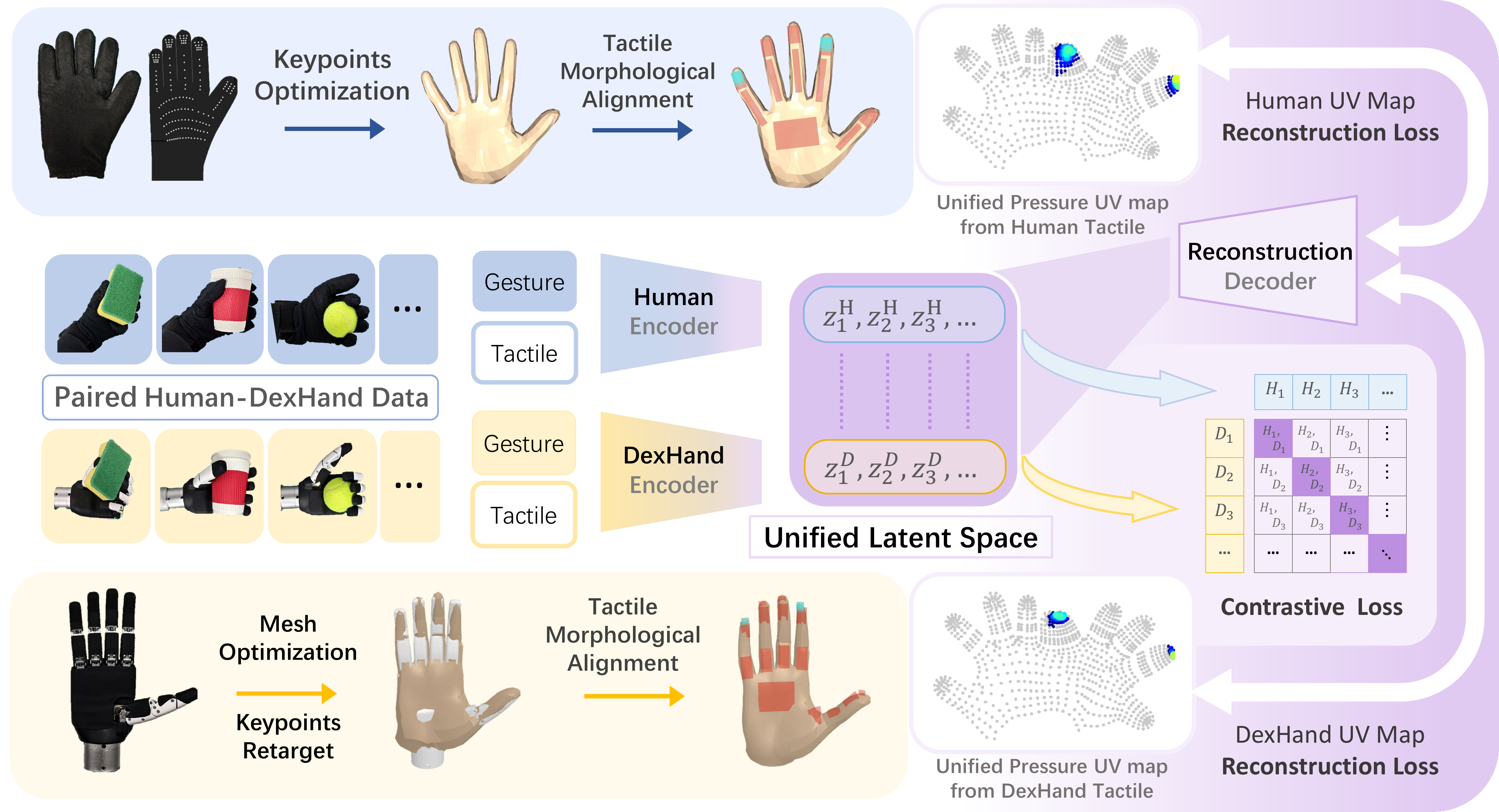}
  \caption{\textbf{An overview of UniTacHand.} (Left) Stage 1: We project tactile data from both human haptic gloves and robotic hands onto a unified MANO UV map. (Right) Stage 2: We introduce contrastive learning with reconstruction and adversarial losses to align the latent representations. We align the tactility and hand gesture from both sources to the same latent space using a contrastive framework trained with paired data. The unified pressure UV maps serve as accurate prior knowledge to supervise the domain-specific encoders, thereby enriching such a latent space with tactile-grounded information.}
  \label{fig:overview}
\end{figure}

To bridge such cross-embodiment gaps, leveraging a shared canonical representation is essential. In the realm of kinematics and vision, the MANO hand model~\citep{MANO} has been established as a cornerstone, providing a parametric space that unifies diverse hand shapes and poses~\citep{handa2020dexpilot}.
MANO model has been widely adopted as a general representation for learning policies on dexterous hands~\citep{wang2023memahand, xie2024ms, crossdex, mandi2025dexmachina, xu2025dexcanvas}. We argue that this geometric consistency can be extended beyond kinematics to unify tactile perception.

Inspired by this, we introduce \textbf{UniTacHand}, a novel unified representation that bridges the tactile gap. Our core idea is to project disparate tactile signals, from human gloves and robotic hands, onto a canonical, morphologically consistent 2D surface space defined by the MANO UV map. Such unification standardizes the heterogeneous data structures and inherently embeds the tactile signals with spatial context.
We utilize a contrastive learning framework to align the two domains into a unified latent space. Based on this unified structure, we demonstrate that 10-minute paired data are sufficient to enable, to our knowledge, the first instance of zero-shot tactile-based policy transfer from human data to a physical robot. Policies trained solely on human data can be directly deployed to perform complex downstream tasks. Furthermore, we demonstrate that co-training on mixed data, including both human and (a small number of) robotic demonstrations via UniTacHand, yields a superior capability. This enables policies to achieve better data efficiency with only one-shot demonstrations on the real robot. UniTacHand provides a general, scalable, and data-efficient pathway for transferring human haptic intelligence to policies on tactile dexterous hands.

Our contributions are as follows: 

\begin{itemize}
    \item We propose \textbf{UniTacHand}, a novel unified spatio-tactile representation that projects heterogeneous tactile data from different hand morphologies to a canonical 2D figure (MANO UV map), and to a unified latent space (by representation learning from paired human-robot data).
    \item We demonstrate zero-shot human-to-robot policy transfer for tactile-based manipulation. Policies trained solely on human data via UniTacHand can be directly deployed on a robot to perform complex tactility-based downstream tasks.
    \item We show that mixing one-shot real robot tactile demonstration with human data provides a superior knowledge for policies, leading to improved performance and data efficiency in terms of human-to-robot policy transfer.
\end{itemize}
\section{Related Work}
\label{sec:related_work}

\paragraph{Canonical hand models as representation space.}
Our work is built upon the foundation of parametric 3D hand models. \citet{loper2015smpl} first introduced SMPL, a learned parametric model for human bodies. Building on this,~\citet{MANO} proposed MANO, a powerful and expressive model of hand shape and articulation derived from thousands of high-resolution 3D scans. MANO has become a standard in the computer vision community, which is widely employed for visual hand pose and shape estimation from images or videos~\citep{boukhayma20193d, zhang2021handoccnet, being0}. Besides, as a kind of motion representation, MANO is also widely used in motion generative models and even vision-language-action (VLA) models~\citep{zhou2025megohand, beingh0, li2025scalable, fu2025handrawer, xie2024ms}. However, its potential as a canonical coordinate frame for tactile data has been largely unexplored. Prior work may project contact onto a mesh~\citep{sundaralingam2023contactmap}. Our work is the first to leverage the intrinsic, morphologically consistent MANO UV map as a unified spatio-tactile representation space specifically for cross-domain human-to-robot policy transfer.

\paragraph{Unified representations for heterogeneous tactile sensors.} The diversity of tactile sensors (e.g., visual-tactile sensors~\citep{yuan2017gelsight, DIGIT}) poses a significant domain adaptation challenge. Early efforts to learn a shared latent space~\citep{li2020connecting}, typically via auto-encoders, discarded the precise spatial geometry essential for manipulation policies. Recent works seek unified representations, but diverge in their target alignment space. UniTouch~\citep{yang2024unitouch} aligns to the semantic space of CLIP~\citep{CLIP} for zero-shot semantic tasks. Some other works including T3~\citep{zhao2024transferable} use physics-based priors (e.g., forces~\citep{wu2024canonical}) or learn data-hungry implicit embeddings, which require millions of samples. On the other hand, paired video-tactility data have been leveraged to align force information with visual contact, extracting knowledge and skills from human demonstrations~\citep{song2025opentouch, xu2025dexcanvas}. \textbf{UniTacHand} presents a more data-efficient and geometrically grounded alternative. We utilize such spatial context as useful signals, and align it to an explicit morphological space. By leveraging the MANO UV map as a powerful inductive bias, our model learns a aligned embedding from a fraction of the data (a 10-minute collected paired set).

\paragraph{Policy transfer from human data.}
Learning from demonstration (LfD) has been well-established for visual or kinematic trajectories~\citep{laskin2020curl, argall2009survey, bidexhd, beingh0}, yet transferring \textit{contact-rich} policies via tactile data remains an open problem. The challenge stems from the embodiment gap -- the profound morphological and sensory mismatch between a human hand and a robotic end effector. For instance, multiple human demonstration datasets are widely used to learn behaviors from human manipulation videos or trajectories~\citep{arctic, h2o, taco, grauman2022ego4d, liu2022hoi4d}, as such datasets provide strong priors of performing manipulation tasks~\citep{zhou2025megohand, beingh0, yang2025egovla}.


In the field of tactile sensing, prior research has often tailored its approach according to the morphological complexity of the target system. For instance, Feel the Force (FTF)~\citep{FTF} achieved compelling zero-shot transfer by unifying representations of contact forces; however, this was demonstrated only on a parallel-jaw gripper, a simple structure that circumvents the challenges of high-DoF, multi-finger coordination. When it comes to more complex, anthropomorphic hands, MimicTouch~\citep{yu2024mimictouch} relies on a sophisticated three-stage pipeline that includes online residual reinforcement learning. This dependence on an online RL patch to correct the policy on the physical robot reveals that the embodiment gap has not been fundamentally addressed at the representation level. In contrast, our work focuses on resolving the embodiment gap by aligning different domains into a canonical space. This approach enables genuine zero-shot transfer for dexterous hands and enriches the learned knowledge by integrating human demonstrations with one-shot data from real robots.
\section{Method}
\label{sec:method}

Our objective is to learn a unified tactile-pose representation that bridges the significant morphological and sensor-level gap between human hands and robotic dexterous hands. Our method, \textbf{UniTacHand}, achieves this through a two-stage process, as illustrated in \cref{fig:overview}. First, we introduce a canonical representation by projecting heterogeneous tactile data from both human and robotic domains to the 2D UV map of the MANO hand model. Such approaches unify the data structure with the same dimensionality and embeds spatial context, making it possible for policies to learn from a mixture of both human and robotics data. Second, we leverage a small, paired dataset to train encoders using a contrastive learning framework, aligning the representations from both domains into a shared, domain-invariant latent space.

\subsection{Unified Representation via MANO UV Maps}
\label{sec:unification}

The MANO model~\citep{MANO} provides a low-dimensional, differentiable parameterization for the human hand, defined by shape parameters $\beta$ and pose parameters $\theta$. Critically, it offers a consistent mesh topology ($N_v=778$ vertices) and a corresponding 2D UV map, which we utilize as our canonical surface space for decoders.

\begin{figure}[t]
    \centering
    \includegraphics[width=0.5\textwidth]{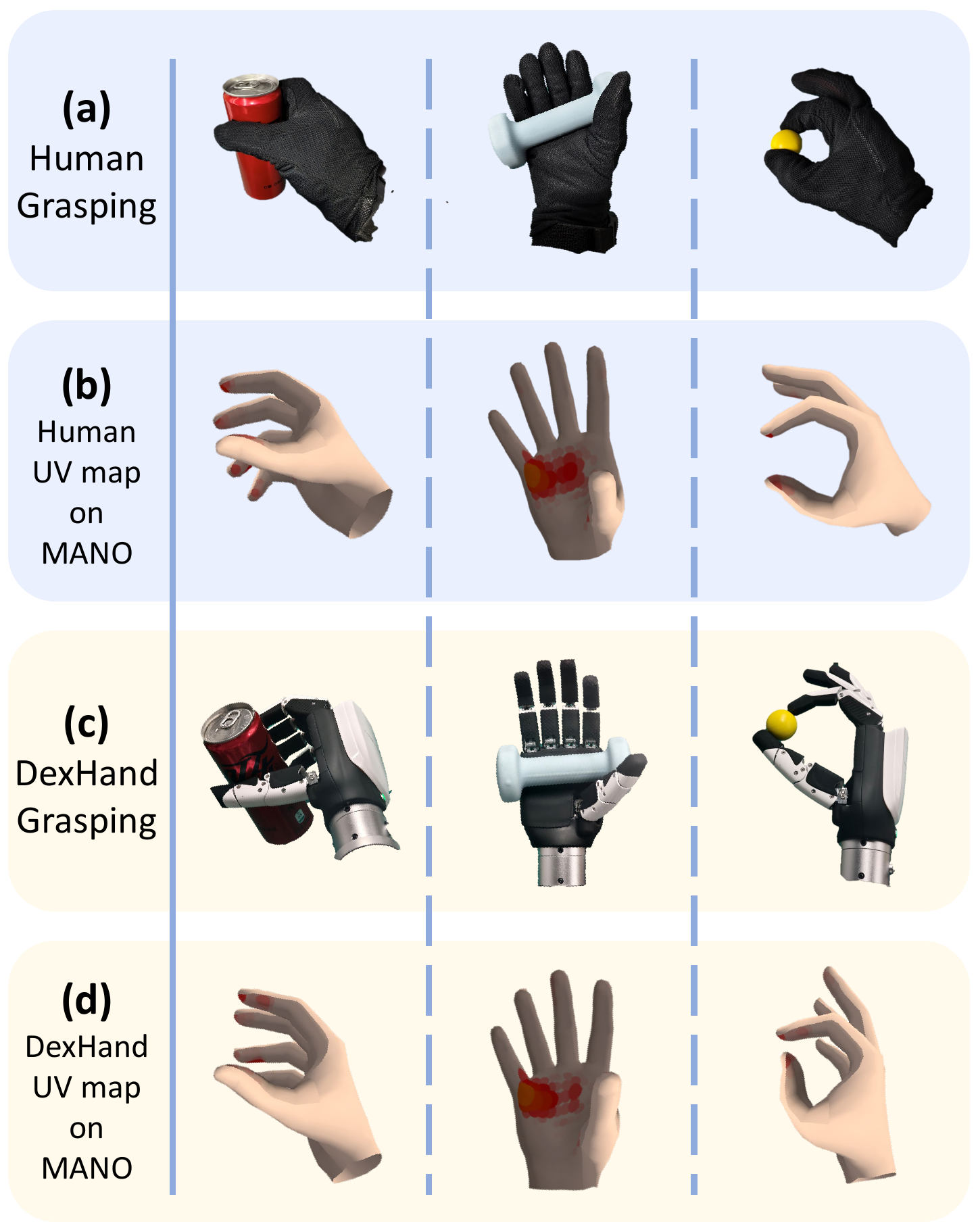} 
    \caption{\textbf{UV mapping results.} When a human hand (or robotic dexterous hand) grasps an object, the activated tactile lattices on a MANO hand are highlighted in red, with the poses of the hand (actions of fingers and rotations of the wrist) rendered at the same time. In this way, we achieve the unification of hand actions and spatial tactility, as well as the alignment between human hands and dexterous hands in terms of tactile information.}
    \label{fig:show_UV}
\end{figure}

\paragraph{Human haptic data projection.}
We acquire human demonstration data using a motion capture glove, which provides 3D keypoints for hand pose, and a pressure-sensitive tactile glove, which yields sparse tactile readings $T_H \in \mathbb{R}^{N_H}$ (where $N_H$ is the number of human sensors). To project these sparse signals onto the continuous MANO surface, we perform a one-time morphological annotation. We manually identify and associate the four corner vertices of each rectangular sensing region on the tactile glove with the corresponding vertex indices on the MANO meshes, which defines a patch on the mesh. The pressure reading from a sensor is then distributed across its associated patch, with values for vertices inside the region computed via bilinear interpolation. This process yields a dense tactile representation on the mesh, which is then rasterized into a tactile UV map $U_H^{ori} \in \mathbb{R}^{W \times H}$.

\paragraph{Robotic hand tactile data projection.}
For the dexterous robotic hand, we leverage its known URDF model and the predefined geometry of its $N_R$ tactile sensors, which provide readings $T_R \in \mathbb{R}^{N_R}$. To bridge the morphological gap to the MANO model, we first perform a one-time optimization to find the optimal MANO shape parameters $\beta^*$. This optimization minimizes a joint loss function $\mathcal{L}_{\text{align}} = \mathcal{L}_{\text{CD}} + w(t) \mathcal{L}_{\text{key}}$   (using a static reference pose for both hands). $\mathcal{L}_{\text{CD}}$ is the Chamfer Distance loss, computed between 4096 points sampled from the robotic hand's URDF mesh and the MANO mesh. $\mathcal{L}_{\text{key}}$ is the keypoint position difference loss based on 21 corresponding keypoints. The keypoint loss ensures correct initial alignment and its weight $w(t)$ is gradually decayed as the optimization proceeds.

Once the optimal shape $\beta^*$ is fixed, we address the dynamic pose. For each frame, we solve a real-time keypoint-based retargeting problem. We directly optimize for the MANO pose parameters $\theta$ that minimize the positional difference between the 21 keypoints derived from the robot's current joint state $P_R$ and the corresponding keypoints on the MANO model (defined by $\beta^*$ and $\theta$). We then project the robot's tactile sensor readings $T_R$ onto this mesh. We model each sensor as a 2D area in the URDF and project this area onto the optimized MANO mesh, transferring the sensor's reading to the corresponding set of mesh vertices using a weighted interpolation scheme. This process generates a corresponding robotic tactile UV map $U_R^{ori} \in \mathbb{R}^{W \times H}$.

\paragraph{Post-processing and masking.}
To ensure spatial continuity and account for minor alignment errors, we first apply a Gaussian smoothing kernel to the populated texels for both $U_H^{ori}$ and $U_R^{ori}$. Let these smoothed maps be $U_H^{smooth}$ and $U_R^{smooth}$.

To focus the learning process only on active sensing regions, we then create binary masks $M_H$ and $M_R$. These masks are $1$ for texels that correspond to a sensor region and $0$ otherwise. The final unified representations, which are used in all subsequent steps, are defined as:
$$
U_H = U_H^{smooth} \odot M_H \quad \text{and} \quad U_R = U_R^{smooth} \odot M_R
$$
where $\odot$ denotes element-wise multiplication.



\begin{figure}[t]
    \centering
    \includegraphics[width=0.68\textwidth]{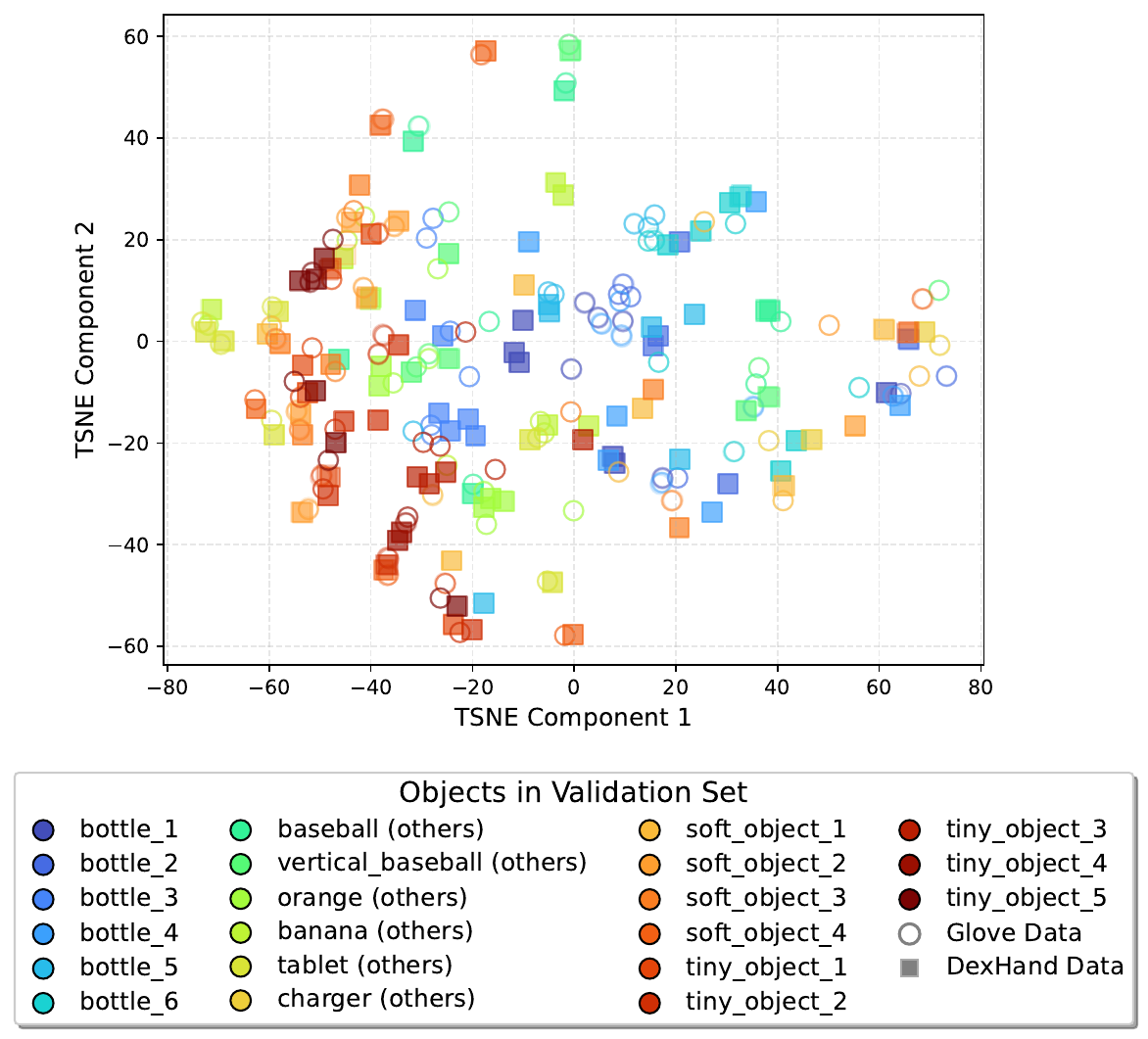} 
    \caption{\textbf{t-SNE visualization of latent space on unseen validation data.} The plot shows latent representations of objects from the validation set (not used in training). Each color represents an object category (e.g., shades of blue for bottle types) and shape indicates the data source (circles: Glove, squares: DexHand). We observe that semantically similar objects, like different bottle or soft\_object samples, cluster closely. This demonstrates that the learned latent space is meaningful and captures generalizable object characteristics.}

    \label{fig:representation}
\end{figure}

\subsection{Latent Space Alignment via Representation Learning}
\label{sec:alignment}
Given the unified tactile UV maps and corresponding pose data, $d_H = (U_H, P_H)$ and $d_R = (U_R, P_R)$, where $P_H$ represents human hand keypoints and $P_R$ is the robot hand's 6-DoF pose and joint state, our goal is to learn two encoders, $E_H$ and $E_R$, that map these inputs into a shared latent space, which is shown in~\cref{fig:representation}.

\paragraph{Architecture.}
Our model consists of two domain-specific encoders, $E_H$ and $E_R$. Each encoder has a two-stream architecture: a CNN backbone processes the high-dimensional tactile UV map ($U_H$ or $U_R$), while a separate MLP processes the low-dimensional pose information ($P_H$ or $P_R$). The features from both streams are then fused to produce a unified latent representation. The latent features are passed to three separate heads: (1) a shared projection head $P$ that maps features into an embedding space for the contrastive loss; (2) two domain-specific decoders, $D_H$ and $D_R$, trained to reconstruct the original tactile UV maps $\hat{U}_H = D_H(E_H(d_H))$ and $\hat{U}_R = D_R(E_R(d_R))$; and (3) a domain classifier $C_D$ for adversarial alignment.

\paragraph{Paired dataset and augmentation.}
We collect a paired dataset of human and robotic hand interactions. This dataset involves manipulating a set of 50 common objects, totaling 688 trajectories of data (16k frames at 40 Hz).

As high-fidelity paired tactile data is challenging to acquire at scale, we focus on data-efficient learning. We introduce a novel augmentation strategy based on the observation that linear interpolations of paired trajectories remain valid pairs. Given two paired data samples from the dataset, $(d_H^i, d_R^i)$ and $(d_H^j, d_R^j)$, we generate a new synthetic pair $(d_H^{new}, d_R^{new})$, where $d_H^{new} = (U_H^{new}, P_H^{new})$ and $d_R^{new} = (U_R^{new}, P_R^{new})$. These new components are generated by interpolating the tactile maps and pose data independently:
\begin{align}
    (U_H^{new}, P_H^{new}) = \left(\lambda_1 U_H^i + (1-\lambda_1) U_H^j,  ~\lambda_2 P_H^i + (1-\lambda_2) P_H^j \right) \\
    (U_R^{new}, P_R^{new}) = \left(\lambda_1 U_R^i + (1-\lambda_1) U_R^j,  ~\lambda_2 P_R^i + (1-\lambda_2) P_R^j \right)
\end{align}

\begin{minipage}{0.46\textwidth}

where $\lambda_1, \lambda_2 \in [0, 1]$ are randomly sampled coefficients to fully utilize all paired information. This interpolation is applied to the full data and, more powerfully, to subsets of fingers (both their motion and tactile data simultaneously). This physics-informed augmentation significantly expands the effective training data. We also apply standard augmentations such as Gaussian noise and sensor dropout. As shown in~\cref{fig:ablation}, the ablation results demonstrate that our method together with the physics-informed augmentation strategy is effective, outperforming a baseline method using contrastive learning on a VAE network or our method without data augmentation.

\paragraph{Training objectives.}
Our model is trained end-to-end with a composite loss function.

\textit{1) Contrastive Alignment Loss:} We use a symmetric InfoNCE loss to align the latent embeddings from paired samples. For a batch of $B$ pairs, let $z_H^i = P(E_H(d_H^i))$ and $z_R^i = P(E_R(d_R^i))$ be the projected embeddings. The loss is:
\end{minipage}%
\hfill
\begin{minipage}{0.5\textwidth}
    \centering
    \includegraphics[width=\linewidth, trim={0cm, 0cm, 0cm, 0cm}, clip]{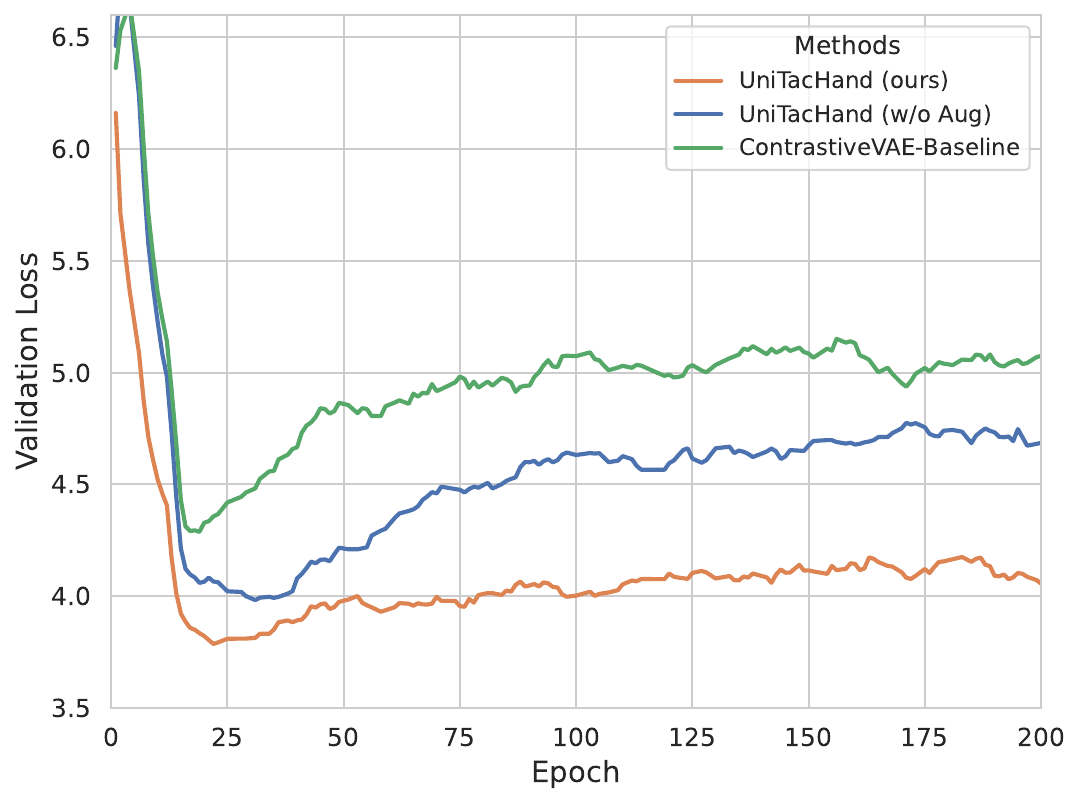}
    \captionof{figure}{\textbf{Ablation study on representation learning.} The validation loss comparison demonstrates that our full UniTacHand framework (orange) outperforms both the ContrastiveVAE Baseline (green) and a variant without our data augmentation strategy (w/o Aug, blue), highlighting the role of augmentation in achieving superior performance.}
    \label{fig:ablation}
\end{minipage}


\begin{equation}
\label{eq:infonce}
\begin{split}
\mathcal{L}_{\text{CON}} = - \frac{1}{B} \sum_{i=1}^{B} \Bigg[\log \frac{\exp(s(z_H^i, z_R^i) / \tau)}{\sum_{j=1}^{B} \exp(s(z_H^i, z_R^j) / \tau)} + \log \frac{\exp(s(z_R^i, z_H^i) / \tau)}{\sum_{j=1}^{B} \exp(s(z_R^i, z_H^j) / \tau)} \Bigg]
\end{split}
\end{equation}
where $s(\cdot, \cdot)$ is the cosine similarity and $\tau$ is a temperature hyperparameter. This loss pulls positive pairs $(z_H^i, z_R^i)$ together while pushing negatives $(z_H^i, z_R^j)$ for $i \neq j$ apart.

\textit{2) Reconstruction Loss:} To ensure the latent space retains high-fidelity information, we task the decoders with reconstructing the original (masked) UV maps. We use a Mean Squared Error (MSE) loss based on the Frobenius norm, computed only over the active sensor regions (as $U_H$ and $U_R'$ are pre-masked):
\begin{equation}
\label{eq:recon}
\begin{split}
\mathcal{L}_{\text{REC}} = \mathbb{E}_{d_H} \left[ || \hat{U}_H - U_H ||_F^2 \right] + \mathbb{E}_{d_R} \left[ || \hat{U}_R - U_R ||_F^2 \right]
\end{split}
\end{equation}

This loss ensures that the encoders do not discard crucial tactile information during the alignment.


\textit{3) Domain-Adversarial Loss:} To forge a truly unified latent space for manipulation, we need to ensure that the features are invariant to their domain (human or robot). So our goal is not merely to align tactile maps, but to create a holistic, manipulation-centric latent space where both action ($P$) and sensation ($U$).

Therefore, we apply a domain classifier $C_D$ to the fused latent representation $E(d)$. The pose inputs ($P_H$ and $P_R$) are heterogeneous, and it is a deliberate design choice. By applying adversarial pressure via a Gradient Reversal Layer (GRL) to the fused representation, we force the encoders to align tactile data and normalize the heterogeneous pose structures into a canonical, domain-invariant representation of hand pose.

This ensures the final latent space $z$ is aligned and agnostic to its origin, which is critical for downstream tasks that depend on the interplay of action and touch. The optimization uses a standard Binary Cross-Entropy (BCE) loss:

\begin{equation}
\label{eq:dann}
\begin{split}
\mathcal{L}_{\text{ADV}} = \mathbb{E}_{d_H} [ \text{BCE}(C_D(E_H(d_H)), 0) ] + \mathbb{E}_{d_R} [ \text{BCE}(C_D(E_R(d_R)), 1) ]
\end{split}
\end{equation}

\begin{minipage}{0.58\textwidth}
    \centering
    \includegraphics[width=1.\linewidth, trim={-3cm, 1cm, -3cm, 0cm}, clip]{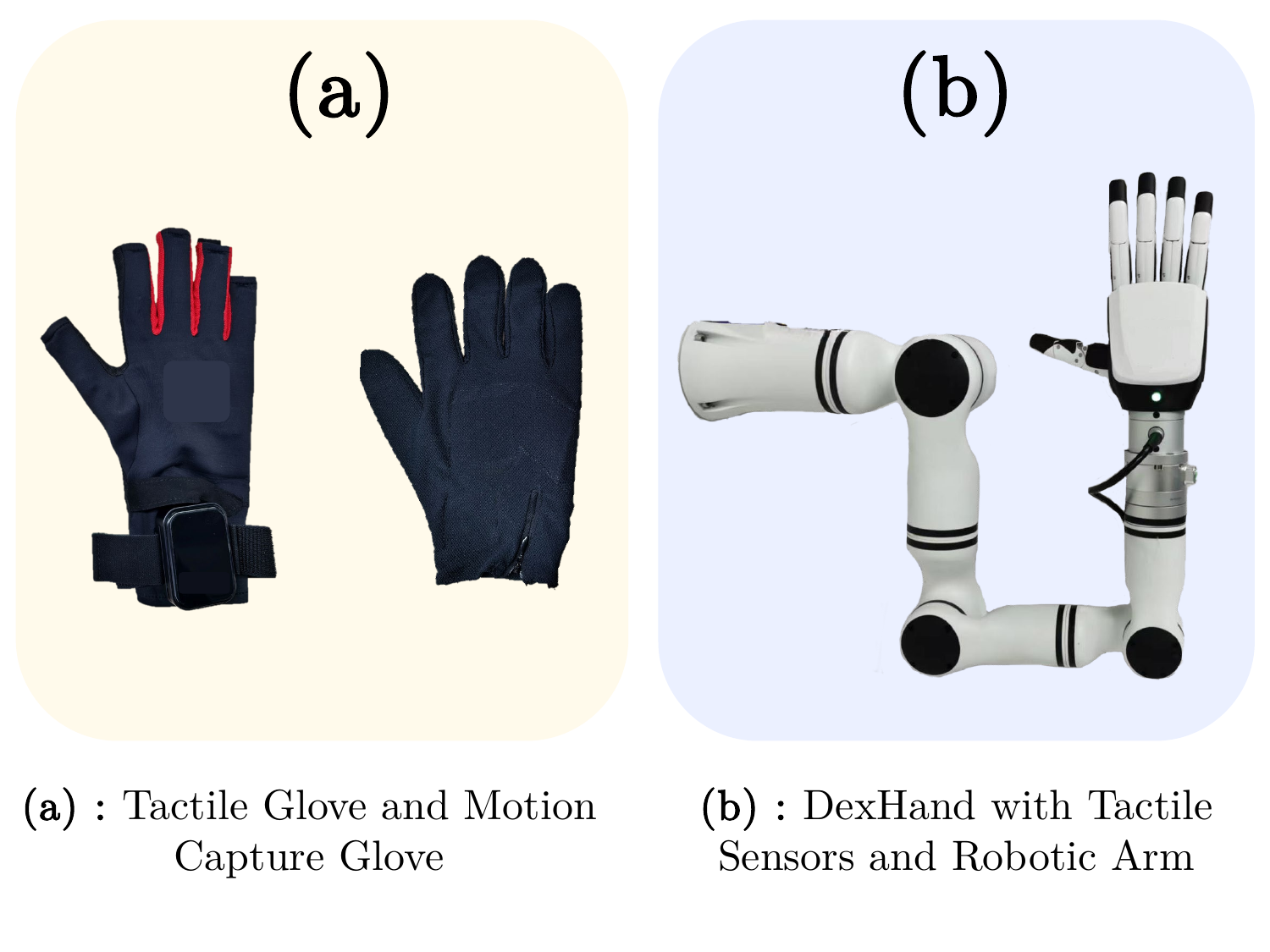}
    \captionof{figure}{\textbf{Hardware settings.} (a) The human contact hardware includes a 137-dim tactile glove with pressure-sensitive fiber sensors on its palm and a motion capture glove. (b) The robotic platform features a 6-DoF RealMan arm and a 6-DoF Inspire tactile hand with 1062-dim arrayed pressure sensors on the surfaces of the fingers and palm.}
    \label{fig:setup}
\end{minipage}
\hfill
\begin{minipage}{0.38\textwidth}
    \centering
    \includegraphics[width=1.1\linewidth, trim={0.5cm, 0cm, 0cm, 0cm}, clip]{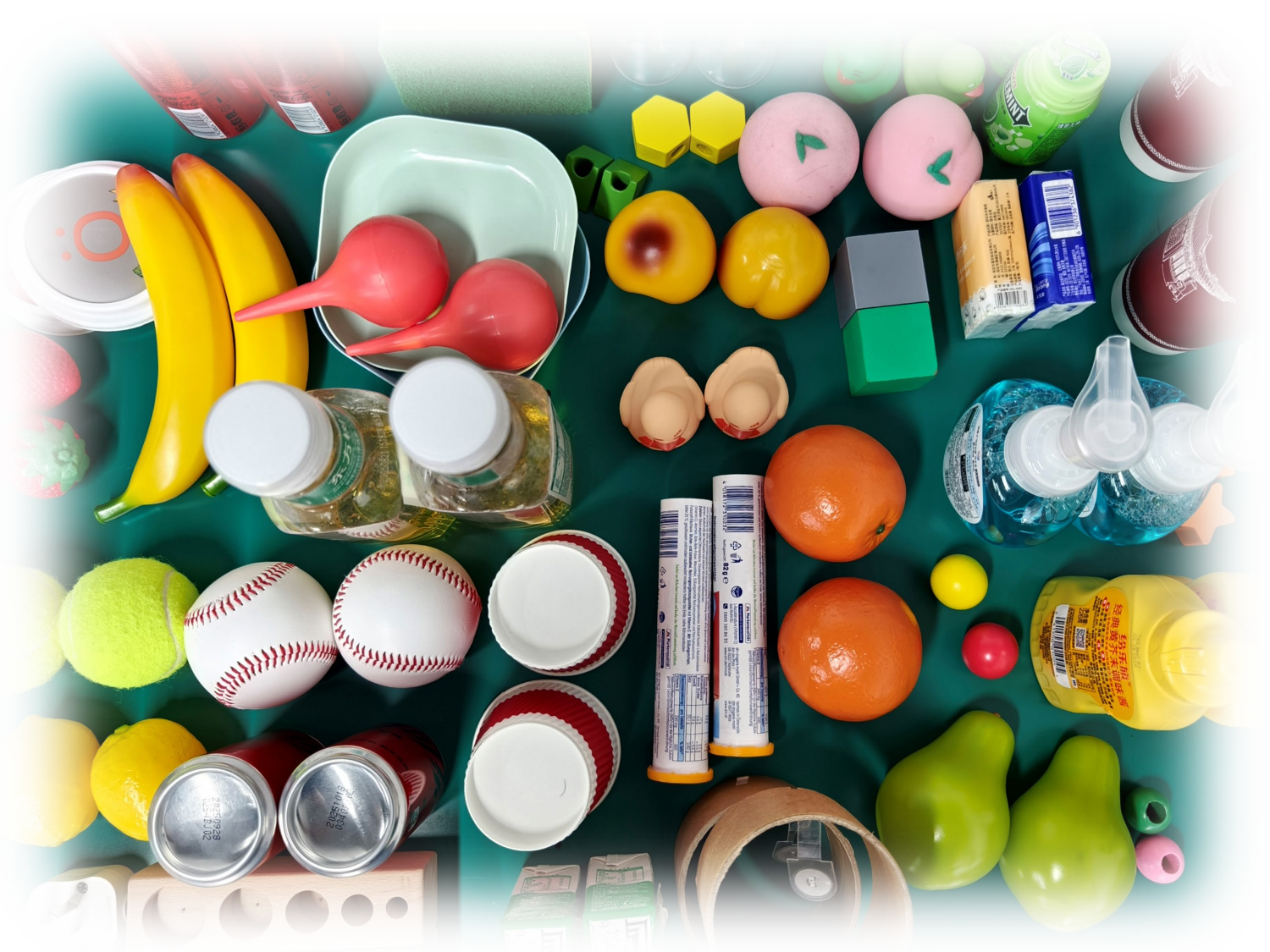}
    \captionof{figure}{\textbf{Examples of the paired dataset.} A subset of our paired object dataset, enabling unified representation alignment by paired manipulation. We choose a variety of objects including fruits, drinks, toys, and other daily items.}
    \label{fig:dataset}
\end{minipage}

\paragraph{Overall objective.}
The final loss function is a weighted sum of these three components:
\begin{equation}
\label{eq:final}
\mathcal{L}_{\text{Total}} = \mathcal{L}_{\text{CON}} + \lambda_{\text{REC}} \mathcal{L}_{\text{REC}} + \lambda_{\text{ADV}} \mathcal{L}_{\text{ADV}}
\end{equation}
where $\lambda_{\text{REC}}$ and $\lambda_{\text{ADV}}$ are hyperparameters balancing the contribution of each term. By optimizing this objective, our encoders $E_H$ and $E_R$ learn to produce representations that are information-rich, and domain-invariant, enabling effective zero-shot and one-shot policy transfer.

\section{Experiments}
\label{sec:experiments}

To validate the effectiveness of \textbf{UniTacHand}, we conduct a series of experiments designed to answer the following questions:

(1) Is our unified spatial-tactile UV Map effective to bridge the domain gap for simple zero-shot tactile tasks?

(2) Through learning from paired human-robot tactile data, can our unified representation be capable of zero-shot human-to-robot policy transfer, with policies trained solely on human data?

(3) By mixing one-shot robot tactile demonstrations with human data, can policies learn from mixed data via \textbf{UniTacHand} show performance gain and data efficiency, demonstrating the capability of performing complex tasks in terms of human-to-robot transfer? 

\subsection{Experimental Setup}

\paragraph{Hardware platform.}
Our experimental setup, shown in~\cref{fig:setup}, consists of a human-side data collection suite and a robot-side deployment platform.
\begin{itemize}
    \item \textbf{Human-side:} We use a custom pressure-sensitive tactile glove providing 137-dimensional tactile readings across the hand. Hand pose is captured by a motion capture glove, which provides 21 keypoint coordinates relative to the wrist.
    \item \textbf{Robot-side:} We use an Inspire tactile hand, which is a 5-finger, 6-DoF anthropomorphic hand equipped with 1062-dimensional tactile sensing capabilities. The hand is mounted on a 6-DoF RealMan robot arm for manipulation tasks. For the robot arm, we use end-effector (EEF) control as the action space, and conduct inverse kinematics (IK) to calculate the target joint angles. For the Inspire hand, the 6 degree-of-freedoms(DoF) naturally corresponds to its 6-dim action space. 
\end{itemize}

\paragraph{Paired dataset for unified alignment in representation learning.}
During representation learning, we train the alignment encoders using a small, 10-minute paired dataset collected from both human and robotic dexterous hands simultaneously~\cref{fig:dataset}. During data collection, we use retargeting algorithms including DexPilot~\citep{handa2020dexpilot} to map MANO keypoints captured from human hand to the Inspire dexterous hand by teleoperation, ensuring that both human and the robot hand are performing the same action in a similar behavioral manner. We collect such paired data on 50 kinds of diverse paired objects , performing the same manipulation process including grasping simultaneously with various hand poses. We collect 688 trajectories containing 16k frames in total, at a frequency of 40 Hz.



\paragraph{Compared methods and baselines.}
We compare our approach against several baselines:
\begin{itemize}
    \item \textbf{Simple patch-wise mapping (PatchMatch):} A baseline that directly matches tactile sensor patches from the human glove to corresponding patches on the robotic hand based on pre-defined spatial correspondence. The lattices on both hands are divided into several partitions, with each paired partition performing a one-to-one mapping.
    \item \textbf{Direct transfer via UV Map (UV-Direct):} A policy is trained on human tactile data projected to the MANO UV map. Such a policy is then directly deployed on the robot hand, using the robotic tactile data projected to the UV map as input.
    \item \textbf{UniTacHand (ours):} Our full method, which uses contrastive learning to align the latent spaces of the human and robot tactile data, and train an encoder-decoder framework in a manner of representation learning. The policy is trained on human data using the human encoder, and transferred to the real robot by using the robot encoder during inference.
\end{itemize}

\begin{table}[t]
  \centering
  \caption{\textbf{Simple zero-shot task evaluation.} Experimental results on \texttt{SoftHardPickPlace} and \texttt{ObjectLocating} tasks. We test each for 20 times and calculate the success rate (\%). Our policies trained using UV Map projection significantly outperform the PatchMatch baselines.}
  \label{tab:simple_tasks}
      \begin{tabular}{ccccc}
        \toprule
        \textbf{Task} & \textbf{PatchMatch} & \textbf{UV-Direct (Ours)} \\
        \midrule
        SoftHardPickPlace & 25.0 & \textbf{85.0} \\
        ObjectLocating & 55.0 & \textbf{100.0} \\
        \bottomrule
      \end{tabular}
\end{table}

\subsection{Unified UV Map for Zero-Shot Policy Transfer}

We first evaluate the fundamental effectiveness of our unified UV map representation on simple, open-loop tactile tasks in a zero-shot setting. In this part, we conduct evaluations on two different tactility-based tasks:

\begin{enumerate}
    \item \textbf{SoftHardPickPlace:} the robot first picks up the object on the table, and places it into the corresponding basket depending on whether it is a soft or hard object. This task requires the ability to distinguish tactile senses according to the softness of the object during contact-rich manipulation.
    \item \textbf{ObjectLocating:} the robot hand touches an object on the tabletop to locate the object according to the tactility, and adjust its own wrist pose following the locating result before finally grasping. In this task, the initial location of the object is randomly sampled within a rectangular range of 10 cm $\times$ 20 cm.
\end{enumerate}

Our policies are trained using only human data and transferred directly to the robot, with results listed in~\cref{tab:simple_tasks}. The experimental results show that our policies trained using UV Map projection significantly outperform the PatchMatch baselines, indicating the effectiveness of our canonical UV Map projection.

\subsection{Unified Representation Learning from Paired Human-Robot Data for Policy Transfer}

\begin{figure}[t]
  \centering
  \includegraphics[width=0.55\textwidth]{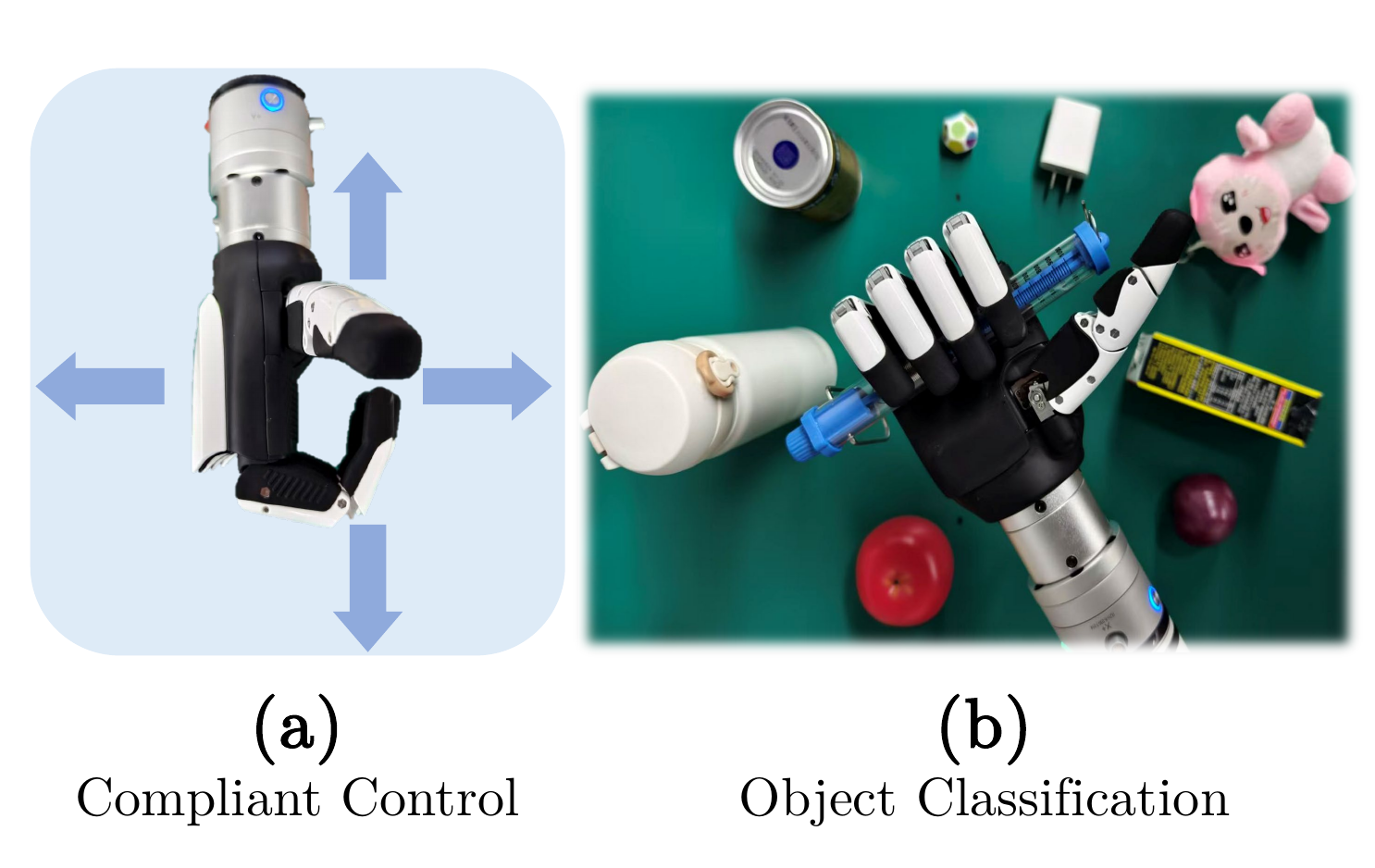} 
  \caption{\textbf{Illustration of the tasks \texttt{CompliantControl} and \texttt{ObjectClassification}.} (a) For \texttt{CompliantControl}, we fix the action of the dexterous hand at a constant pose, and drag parts of the hand to apply external forces on the tactile sensors. The aim of the task is to make the end effector move towards the same direction as the external forces. (b) For \texttt{ObjectClassification}, we choose 10 objects that are totally unseen in the paired data used to train the representation encoders. The aim of the task is to classify the correct category of the objects according to tactile information when grasped in hand.}
  \label{fig:objects}
\end{figure}

To evaluate our unified representation learning framework in terms of human-to-robot policy transfer, we further investigate two more complex zero-shot transfer tasks (as shown in~\cref{fig:objects}):

\begin{enumerate}
    \item \textbf{CompliantControl:} while the robot hand holds a fixed action, the end-effector of the robot arm should follow the human when external forces are applied to the dexterous hand. Such a task requires the capability of spatial perception related to the direction of the external force, which ultimately determines the moving direction of the end-effector. This task is considered successful only when the moving direction inferred from the policy is consistent with the external force applied by human.
    \item \textbf{ObjectClassification:} we choose 10 objects different in shape, textures, geometric features, as well as physical characteristics, to form an object set for the classification task. Holding the object in hand and taking the corresponding tactile information, the policy should infer the category of the object. These 10 objects contain both seen objects in the paired data used for representation learning, and unseen ones.
\end{enumerate}



\begin{table*}[t]
    \renewcommand{\arraystretch}{1.15}
    \caption{\textbf{Detailed results of the tasks \texttt{CompliantControl} and \texttt{ObjectClassification}.} We calculate the accuracy (\%) for each task on each method.}
    \label{tab:complex_discrimination}
    \vspace{2mm}
    \begin{center}
        \begin{tabular}{ccccc}
            \toprule
            \multicolumn{2}{c}{\textbf{Tasks}} & \multicolumn{1}{c}{\textbf{PatchMatch}} & \multicolumn{1}{c}{\textbf{UV-Direct}} & \multicolumn{1}{c}{\textbf{UniTacHand (Ours)}} \\
            \midrule
            \multicolumn{2}{c}{CompliantControl} & \multicolumn{1}{c}{10.0} & \multicolumn{1}{c}{36.0} & \multicolumn{1}{c}{\textbf{40.0}} \\
            \cline{1-5}
            ~ & Human Validation Set & 43.2 & \textbf{71.6} & 59.5 \\
            \cline{2-5}
            {\multirow{-2}*{\makecell[c]{ObjectClassification}}} & Real Robot Test Set  & 15.7 & 18.9 & \textbf{38.6} \\
            \bottomrule
        \end{tabular}
    \end{center}
\end{table*}

\begin{table}[t]
    \renewcommand{\arraystretch}{1.15}
    \caption{\textbf{One-shot manipulation task success rate (\%).} We compare our methods with both visual-only baselines and methods that are solely trained on real robot data.}
    \label{tab:manipulation_tasks}
    \vspace{2mm}
    \begin{center}
        \begin{tabular}{ccc}
            \toprule
            \multicolumn{2}{c}{\textbf{Methods}} & \multicolumn{1}{c}{\textbf{Success Rate}} \\
            \midrule
            ~ & \textbf{R-Visual-Only} & 43.3 \\
            \cline{2-3}
            {\multirow{-2}*{\makecell[c]{Robot Data Only}}} & \textbf{R-Visual-Tactile}  & 50.0 \\
            \cline{1-3}
            ~ & \textbf{PatchMatch} & 56.7 \\
            \cline{2-3}
            ~ & \textbf{UV-Direct} & 63.3 \\
            \cline{2-3}
            {\multirow{-3}*{\makecell[c]{Human Data w/ \\ One-Shot Robot Data}}} & \textbf{UniTacHand (Ours)} & \textbf{73.3} \\
            \bottomrule
        \end{tabular}
    \end{center}
\end{table}

We compare the performances of \textbf{UniTacHand}, \textbf{UV-Direct}, as well as the \textbf{PatchMatch} baselines in~\cref{tab:complex_discrimination}. For the task \texttt{CompliantControl}, we test in five different directions for 50 times and calculate the accuracy (consistency with the direction of the external force). For the task \texttt{ObjectClassification}, we evaluate the classification accuracy on both a held-out human validation set and on the physical robot test set (for zero-shot transfer).

According to the experimental results, due to the cluttered data of human tactile demonstrations, simply applying one-to-one PatchMatch does not work on the task \texttt{CompliantControl}, while our UniTacHand manages to overcome the difficulty of distribution shift from human data to real robot deployment.
Besides, according to the task \texttt{ObjectClassification}, the results clearly demonstrate that while direct UV map projection is a useful representation (which yields a relatively high accuracy in human validation set), it is insufficient to bridge the domain gap, especially due to the different manipulation poses, gestures, and the morphological mismatch in human-robot pairs (a relatively low accuracy in the real robot test set). In this case, our contrastive alignment framework (UniTacHand) successfully learns a shared embedding space, enabling robust zero-shot transfer of a complex classification policy.

Overall, evaluations on both tasks demonstrate that with a more precise and physically consistent representation learned from paired human-robot data, not only do our policies gain better performances, but also they are more capable of generalization in the face of unseen objects and out-of-distribution cases. 

\subsection{One-Shot Learning for Human-to-Robot Policy Transfer}

Finally, we evaluate the ability of our UniTacHand when a one-shot robot demonstration is given. We consider a task with multi-modal inputs: both vision and tactility. In this task, we choose three pairs of objects that share similar appearances, but result in different distributions of tactility due to their physical attributes. For instance, a soft orange and a hard one, an empty bottle and the exact same one full of water -- each pair of objects can be similar in vision, but can be distinct in tactility when being held in hand.


We evaluate our methods and the \textbf{PatchMatch} baseline on this task, together with visual-only policies (\textbf{R-Visual-Only}) as well as policies that only learn from real robot data (\textbf{R-Visual-Tactile}). As listed in~\cref{tab:manipulation_tasks}, our methods strongly outperform various baselines. The results that \textbf{R-Visual-Only} performs worse than methods that utilize tactile information as inputs indicate that in such cases that visual information often makes policy confusing, adding tactility can help the policy distinguish and make feasible decisions. On the other hand, the results that \textbf{R-Visual-Tactile} performs no better than those trained on mixed data (human data with one-shot real robot data) demonstrate the effectiveness of human data in the domain of human-to-robot policy transfer. Despite the fact that human data cannot be directly applied during real robot deployment, such rich and diverse information still provides essential knowledge and helps the process of policy learning.





\section{Conclusion}

In this paper, we propose \textbf{UniTacHand} to unify tactile representations between human hands and dexterous robotic hands. By mapping heterogeneous tactile information from different embodiments onto MANO UV maps and using contrastive learning, our method achieves either zero-shot or one-shot human-to-robot policy transfer for tactile-based manipulation tasks. 
Such unified representations, serving as precise and physically consistent mappings between human hands and dexterous hands, make it possible to utilize large-scale tactile data collected from humans, which is much easier to obtain compared to robot data. 
Future work could scale human tactile data collection and incorporate tactile modality into multi-modal foundation models, such as vision–language–action models, to build generalist robotic systems.

\bibliographystyle{plainnat}
\bibliography{ref}

\clearpage
\appendix

\section{Implementation Details}
\label{sec:appendix_impl}

This appendix provides comprehensive details regarding data processing, network architectures, and training procedures utilized in our \textbf{UniTacHand} framework, supplementing the methodology described in \cref{sec:method}.

\subsection{Data Unification and Preprocessing}

\paragraph{MANO Model Configuration.}
We utilize the standard MANO model \citep{MANO} featuring $N_v=778$ vertices and $N_f=1538$ faces. All tactile projections are rasterized into a normalized 2D UV map with a resolution of $W \times H = 1024 \times 1024$.

\paragraph{Human Data Projection.}
The morphological alignment between the tactile glove and the MANO mesh is performed via a one-time manual annotation. The glove contains $N_H = 137$ sensor arrays. We associate the four corners of each sensor's rectangular region with corresponding vertex indices on the MANO mesh. During processing, pressure values are distributed to vertices within these patches using bilinear interpolation to ensure smooth gradients.

\paragraph{Robotic Hand Alignment.}
To bridge the morphological gap, we perform a two-stage alignment process. First, an offline shape optimization is conducted to determine the optimal MANO shape parameters $\beta^*$ by minimizing the alignment loss over 10000 iterations:
\begin{itemize}
    \item \textbf{Chamfer Distance ($\mathcal{L}_{\text{CD}}$):} Computed between 4096 points uniformly sampled from the robot's URDF mesh surface and 4096 points from the MANO mesh.
    \item \textbf{Keypoint Loss ($\mathcal{L}_{\text{key}}$):} Defined as the $L_2$ distance between 21 semantically corresponding keypoints on both hands.
    \item \textbf{Loss Scheduling:} The weight $w(t)$ for $\mathcal{L}_{\text{key}}$ follows a linear decay schedule $w(t) = \max(0, 1.0 - t / 2500)$, prioritizing global structural alignment in early iterations and surface detail refinement in later stages.
\end{itemize}
Second, for online processing, we employ a real-time pose retargeting module. We optimize the MANO pose parameters $\theta$ frame-by-frame using the Adam optimizer to minimize the keypoint discrepancies between the robot's current joint state and the MANO model.

\paragraph{Post-processing.}
To mitigate rasterization artifacts, a Gaussian smoothing kernel (size $5 \times 5$, $\sigma = 0.5$) is applied to the initial UV maps $U_H^{ori}$ and $U_R^{ori}$. The resulting maps are masked by binary validity masks $M_H$ and $M_R$ and normalized to $[0, 1]$ before network input.

\subsection{Network Architecture}

The UniTacHand architecture adopts a partitioned, region-aware design to handle the morphological discrepancies between the human glove and the robotic hand. It comprises domain-specific encoders ($E_H$, $E_R$) for tactile and pose modalities, followed by a shared latent fusion module.

\paragraph{Tactile Encoders ($E_H^{tac}$, $E_R^{tac}$).}
Unlike standard monolithic encoders, we propose a partition-based encoding strategy that processes specific morphological regions independently before global aggregation.

\begin{itemize}
    \item \textbf{Human Tactile Encoder ($E_H^{tac}$):} The 137-dimensional glove input is divided into 7 semantic regions: \textit{thumb, index, middle, ring, pinky, palm,} and \textit{bend}. Each region $\mathcal{R}_i$ is processed by a specific lightweight MLP branch.
    \begin{itemize}
        \item \textit{Branch Architecture:} The finger branches utilize a 2-layer MLP (hidden dims $[8, 8]$ and $[8, 4]$), while the palm branch uses a deeper structure $[16, 8, 8]$ to capture complex contact patterns.
        \item \textit{Aggregation:} The features from all branches are concatenated and passed through a global mixing MLP with hidden units $[128]$ and an output dimension of $64$ ($D_{tac}$).
    \end{itemize}

    \item \textbf{Robot Tactile Encoder ($E_R^{tac}$):} The 1062-dimensional tactile array is segmented into 17 distinct spatial patches (e.g., \textit{finger\_phalanges, palm}). Each patch is treated as a local image and processed by a dedicated CNN branch selected based on the patch resolution:
    \begin{itemize}
        \item \textit{Small Patches ($3\times3$):} Processed by a single convolutional layer ($C_{out}=1, k=2, s=1$).
        \item \textit{Standard Patches ($12\times8, 10\times8$):} Processed by a 2-layer CNN block. Layer 1: ($C_{out}=1, k=3, s=2$); Layer 2: ($C_{out}=2, k=2, s=1$).
        \item \textit{Large Patches (Palm $8\times14$):} Processed by a 2-layer CNN block with a larger receptive field. Layer 1: ($C_{out}=1, k=4, s=2$); Layer 2: ($C_{out}=2, k=2, s=1$).
        \item \textit{Aggregation:} Feature maps from all branches undergo adaptive average pooling to a fixed $4\times4$ spatial size, are flattened, concatenated, and fused via a global mixing MLP (hidden $[128]$, output $64$).
    \end{itemize}
\end{itemize}

\paragraph{Pose Encoders ($E_H^{pose}$, $E_R^{pose}$).}
Pose information is encoded via multi-layer perceptrons to align the kinematic spaces.
\begin{itemize}
    \item \textbf{Human Side:} The 60-dimensional input is processed by a 4-layer MLP with hidden dimensions $[64, 32, 64]$ and a final projection to $32$ ($D_{pose}$).
    \item \textbf{Robot Side:} The 6-dimensional joint angles are processed by a 3-layer MLP with hidden dimensions $[32, 64]$ and a final projection to $32$ ($D_{pose}$).
\end{itemize}

\paragraph{Fusion and Projection.}
The extracted tactile and pose features are concatenated ($D_{fused} = 32 + 64 = 96$) and mapped to the shared latent space.
\begin{itemize}
    \item \textbf{Post-Concat Fusion:} A transition MLP with hidden size $128$ reduces the dimensionality to $64$.
    \item \textbf{Shared Projection Head:} To obtain the final unified representation $z \in \mathbb{R}^{32}$, the features pass through a shared MLP block with hidden dimensions $[128, 128]$.
\end{itemize}

\paragraph{Decoders and Discriminator.}
To validate the representation quality, we employ symmetric decoders ($D_H, D_R$) to reconstruct the UV representations from the shared latent space $z$.
Unlike the partition-based encoders, the decoders utilize a global Multi-Layer Perceptron (MLP) architecture.
Specifically, the network expands the latent vector through hidden layers with dimensions $[128, 256]$ and projects it to a final output dimension of $391$, corresponding to the UV feature space for both the human glove and the Inspire hand.
The domain discriminator $C_D$ retains a 3-layer MLP structure $[128, 64, 32, 16, 1]$ with LeakyReLU activations, receiving input via a Gradient Reversal Layer (GRL) to enforce domain invariance.

\subsection{Training Protocol}

\paragraph{Dataset Split.}
The paired dataset (688 trajectories, $\sim$16k frames) is split into training and validation sets with a ratio of 70\%:30\%. Splits are stratified by object instance to ensure zero-shot evaluation on unseen objects.

\paragraph{Optimization.}
We train end-to-end using the Adam optimizer ($lr=5e-5$, weight decay=$1e-5$) with a batch size of $B=1024$ for 200 epochs. 

\paragraph{Loss Configuration.}
The total loss weights are set to $\lambda_{\text{REC}} = 1.0$ and $\lambda_{\text{ADV}} = 0.5$. The InfoNCE temperature is $\tau = 0.1$.

\paragraph{Augmentation Details.}
Our proposed interpolation augmentation samples mixing coefficients $\lambda_1, \lambda_2 \sim U(0, 1)$. Standard augmentations include additive Gaussian noise ($\sigma=0.03$) and random sensor dropout (probability $p=0.005$).

\section{Ablation Study}
\label{sec:ablation}

To validate the design choices of UniTacHand, we conduct extensive ablation studies focusing on data augmentation strategies and loss components.

\subsection{Impact of Augmentation Strategies}

One key contribution of our method is the physics-informed interpolation augmentation. To rigorously quantify its impact and disentangle it from standard regularization techniques, we trained two additional variants of our model:
\begin{itemize}
    \item \textbf{w/o Linear Aug (Red):} Removes only our proposed paired interpolation strategy.
    \item \textbf{w/o Aug (Blue):} Removes all data augmentations, including Gaussian blur and dropout.
\end{itemize}

\begin{figure}[t]
    \centering
    \includegraphics[width=0.6\linewidth]{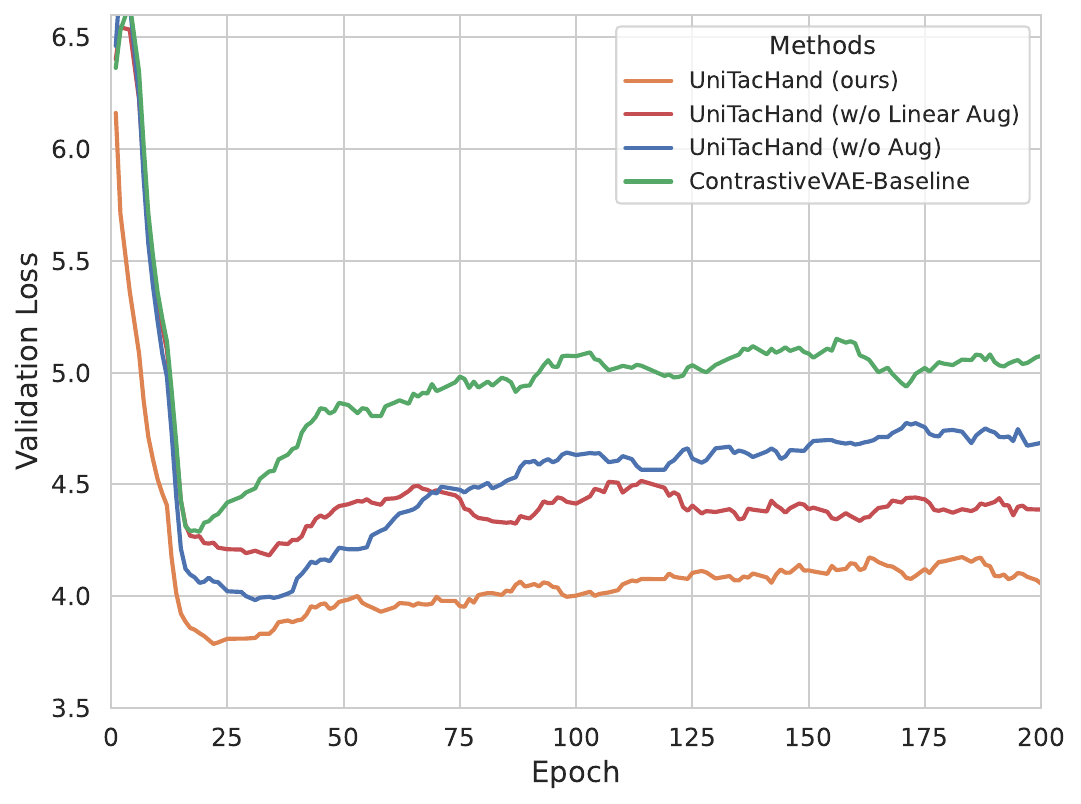}
    \caption{\textbf{Ablation on augmentation strategy.} Comparison of validation loss curves. We compare our full method (Orange) against a variant without our specific linear interpolation (Red) and a baseline without any augmentation (Blue), demonstrating the additive value of our proposed strategy.}
    \label{fig:ablation_aug}
\end{figure}

As shown in \cref{fig:ablation_aug}, the model trained with our full augmentation strategy (Orange curve) exhibits the fastest convergence and the lowest final validation loss. 

Comparing the variants, removing all augmentations (Blue curve) results in the highest loss. However, simply adding these standard augmentations is not enough; the gap between the red curve (w/o Linear Aug) and the orange curve (Ours) demonstrates that our physics-informed interpolation effectively densifies the training manifold, significantly boosting generalization to unseen contact configurations beyond what standard noise injection can achieve.

\subsection{Impact of Decoder Configurations}
\label{sec:ablation_decoder}

We further investigate the effectiveness of our latent representation learning by comparing different decoder configurations. Specifically, we evaluate our default \textbf{Dual Decoder} against single-branch variants (\textbf{DexHand-only} and \textbf{Human-only}) and a baseline that directly reconstructs \textbf{Raw Tactile} signals.

\begin{figure}[t]
    \centering
    \includegraphics[width=0.6\linewidth]{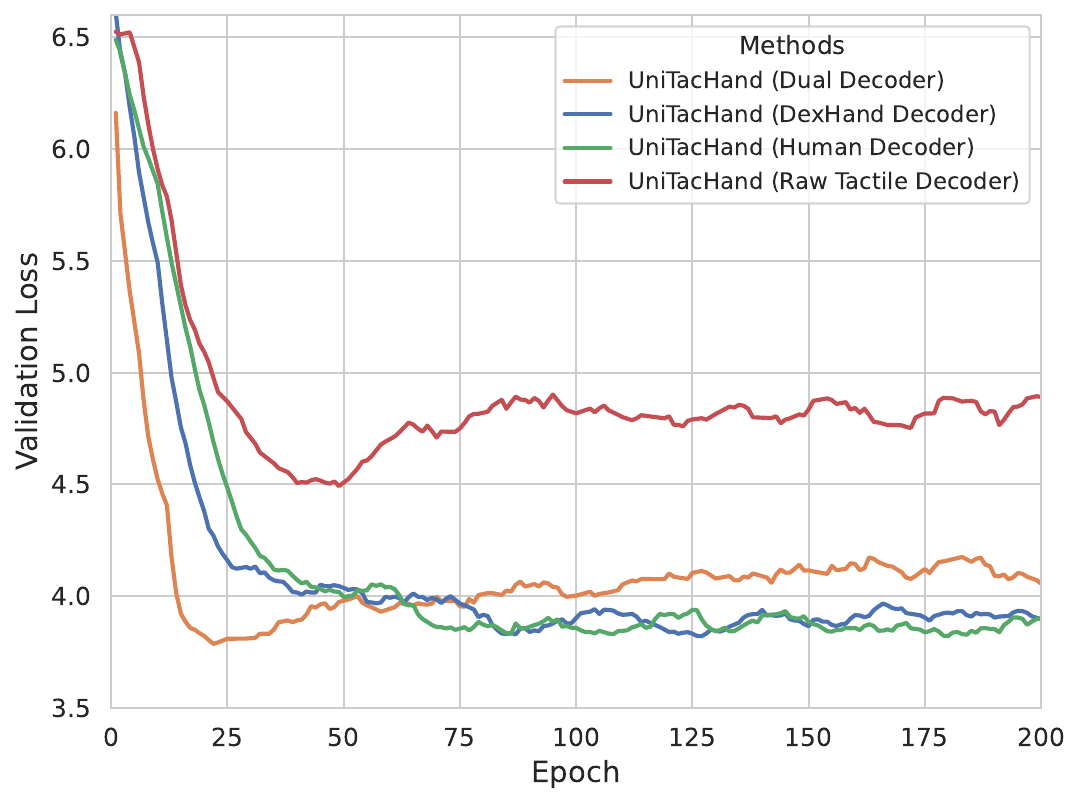} 
    \caption{\textbf{Comparison of decoder architectures.} The Dual Decoder (Orange) achieves the fastest initial convergence, while the Raw Tactile baseline (Red) performs the worst due to signal noise. The subsequent rise in the Orange curve indicates overfitting due to limited paired data capacity.}
    \label{fig:ablation_decoder}
\end{figure}

As illustrated in \cref{fig:ablation_decoder}, the Raw Tactile Decoder (Red curve) yields the highest validation loss throughout the training. We attribute this to the high dimensionality and noisy details inherent in raw sensor data, which make it challenging for the model to extract meaningful semantic representations compared to the structured geometry of UV maps.

In contrast, all UV-based decoders achieve significantly lower losses, confirming that both Human and DexHand UV maps provide effective supervision signals. Notably, the \textbf{Dual Decoder (Orange)} exhibits the fastest convergence rate in the early stages, suggesting that leveraging constraints from both modalities simultaneously accelerates feature learning. 

However, we also observe that the Dual Decoder begins to overfit after approximately 25 epochs (loss increases), whereas the single-decoder variants (Blue and Green) maintain stable convergence. We hypothesize that while the dual-objective provides stronger gradients initially, our current volume of paired data may be insufficient to support the simultaneous high-fidelity reconstruction of two complex modalities without overfitting. Consequently, while the Dual setup is powerful, single decoders offer a trade-off with slightly slower convergence but greater stability on this dataset.

\subsection{Alignment Visualization}
\label{sec:ablation_viz}

\begin{figure}[t]
    \centering
    \begin{subfigure}[b]{0.32\linewidth}
        \centering
        \includegraphics[width=\linewidth]{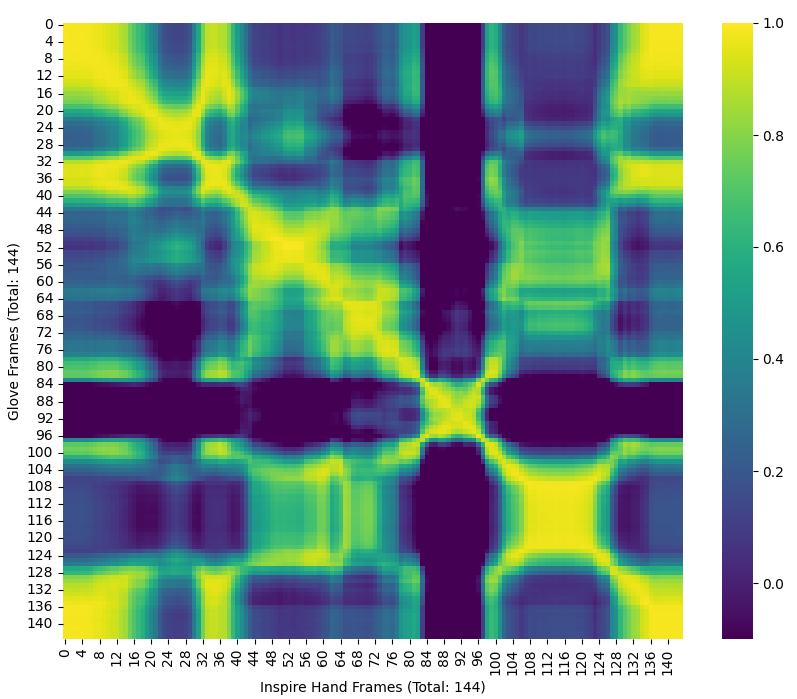}
        \caption{Ours}
        \label{fig:hm_ours}
    \end{subfigure}
    \hfill
    \begin{subfigure}[b]{0.32\linewidth}
        \centering
        \includegraphics[width=\linewidth]{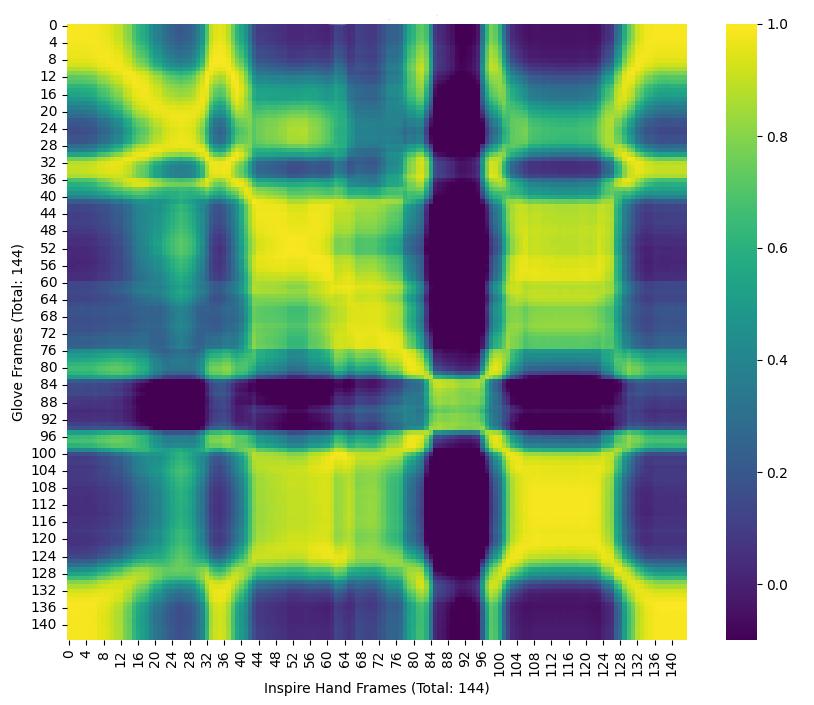}
        \caption{w/o UV Decoder}
        \label{fig:hm_no_uv}
    \end{subfigure}
    \hfill
    \begin{subfigure}[b]{0.32\linewidth}
        \centering
        \includegraphics[width=\linewidth]{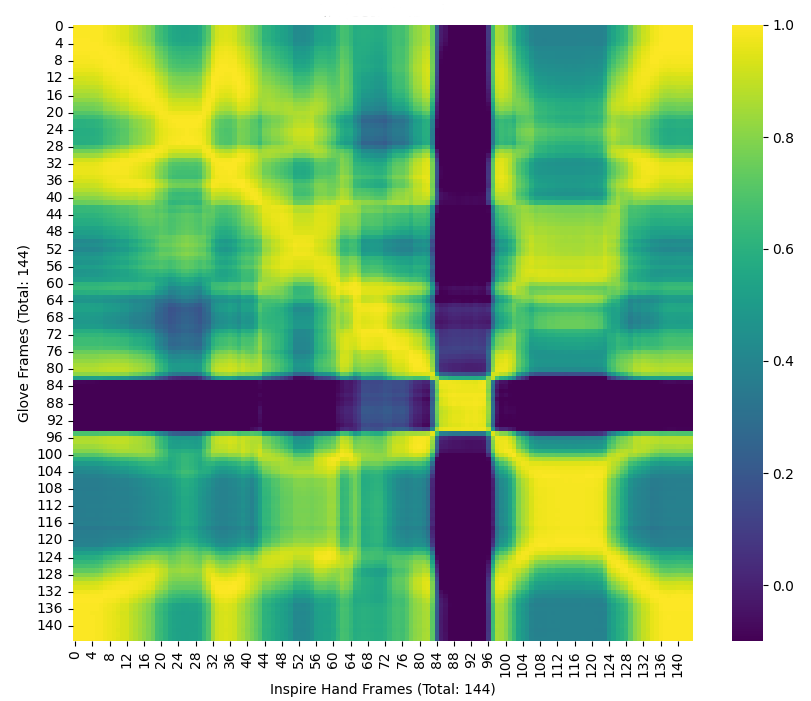}
        \caption{w/o Augmentation}
        \label{fig:hm_no_aug}
    \end{subfigure}
    \caption{\textbf{Latent similarity heatmaps.} Comparison of feature alignment quality on a paired validation trajectory. (a) \textbf{Ours} shows a distinct diagonal with clear boundaries. (b) Removing the UV decoder leads to local ambiguities (small blocky confusion). (c) Removing augmentation results in widespread confusion off the diagonal.}
    \label{fig:heatmap_vis}
\end{figure}

To qualitatively assess the alignment quality, we visualize the cosine similarity matrix of latent representations for a paired trajectory from the validation set. As shown in \cref{fig:heatmap_vis}, the $x$ and $y$ axes represent the time steps of the Human and Robot tactile sequences, respectively. An ideal alignment should manifest as a sharp, high-similarity diagonal.

Comparing the visualizations reveals distinct behaviors:
\begin{itemize}
    \item \textbf{Ours (\cref{fig:hm_ours}):} Displays a crisp diagonal line with a clean background. This indicates that our method successfully learns a temporally consistent and distinct representation, where specific contact states are uniquely matched across modalities.
    \item \textbf{w/o UV Decoder (\cref{fig:hm_no_uv}):} Exhibits minor block-like confusion around the diagonal. Without the dense geometric supervision from the UV map, the model struggles to differentiate between spatially adjacent tactile patterns, leading to local aliasing.
    \item \textbf{w/o Augmentation (\cref{fig:hm_no_aug}):} Suffers from large-scale confusion areas (high similarity in off-diagonal regions). This confirms that without our physics-informed augmentation, the model overfits to specific signal amplitudes rather than learning robust contact features, causing it to confuse temporally distant frames.
\end{itemize}

\section{Tasks and Policy Training}

In this section, we describe the details of the task definitions, including the basic settings of observation and action spaces, the location of the objects, as well as the success conditions. We also present the details of training policies for these tasks.

\begin{figure}[t]
    \centering
    \includegraphics[width=0.48\textwidth]{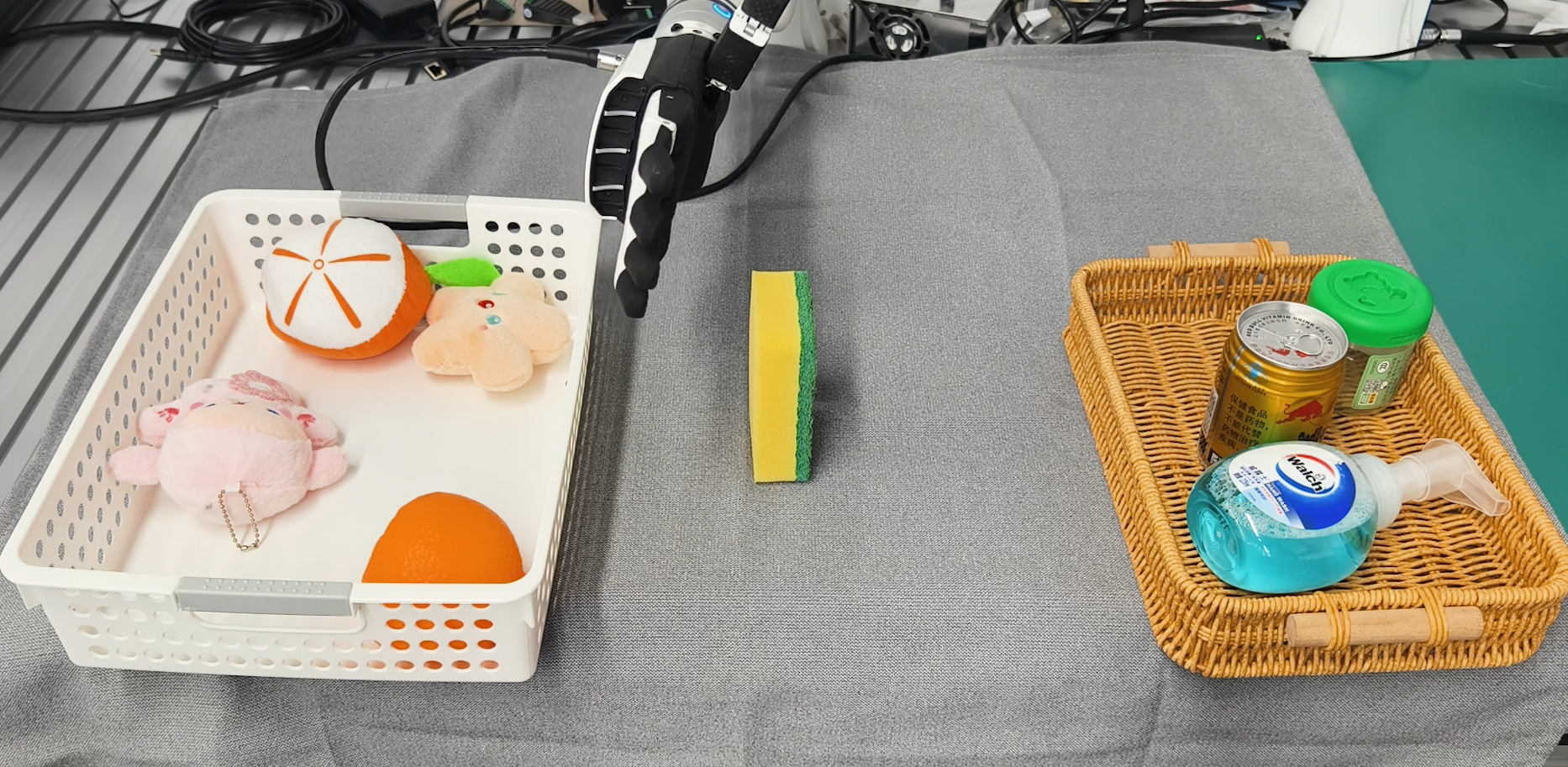} 
    \caption{\textbf{An example of the task \texttt{SoftHardPickPlace}.} The robot picks up the object, and places it into one of the two baskets according to whether the object is soft or hard.}
    \label{fig:appendix-task1}
\end{figure}

\begin{figure}[t]
    \centering
    \includegraphics[width=0.48\textwidth]{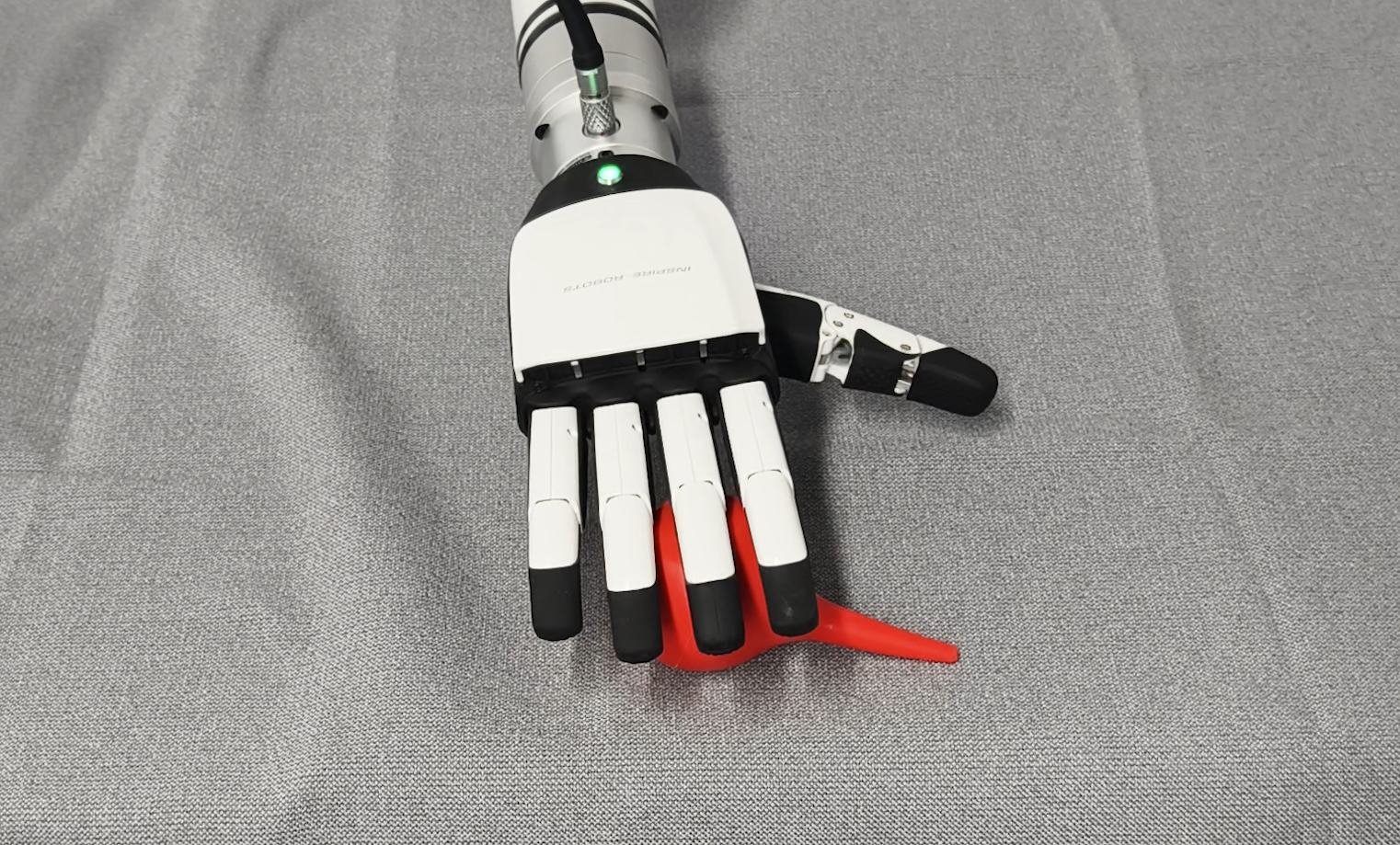} 
    \caption{\textbf{An example of the task \texttt{ObjectLocating}.} The robot touches the object first, then adjusting the movement of the end-effector, so as to grasp the object successfully.}
    \label{fig:appendix-task2}
\end{figure}

\begin{figure}[t]
    \centering
    \includegraphics[width=0.48\textwidth]{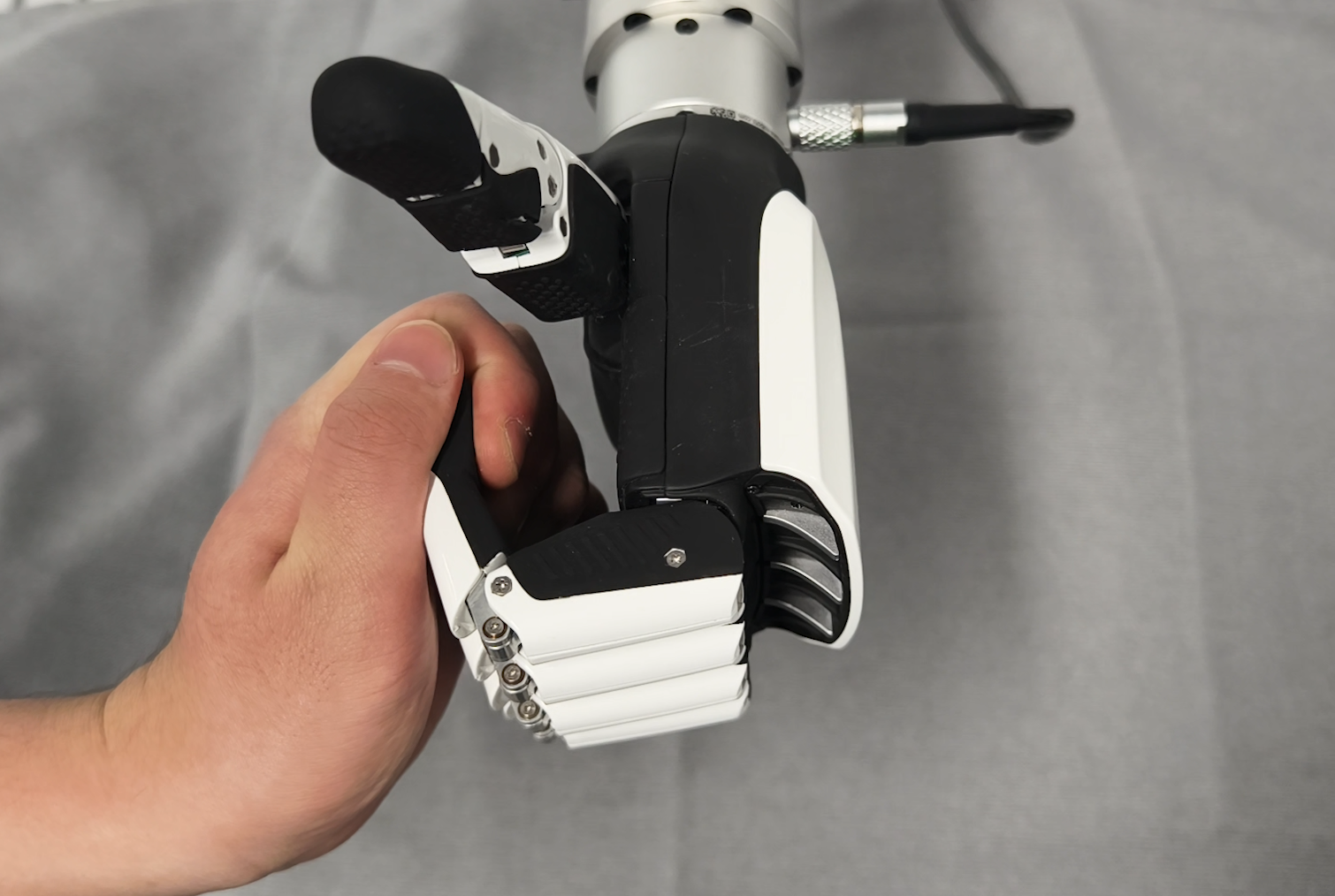} 
    \caption{\textbf{An example of the task \texttt{CompliantControl}.} A person grabs the dexterous hand by applying external forces to a certain direction. The policy should infer the correct direction and move the end-effector according to tactility.}
    \label{fig:appendix-task3}
\end{figure}

\begin{figure}[t]
    \centering
    \includegraphics[width=0.48\textwidth]{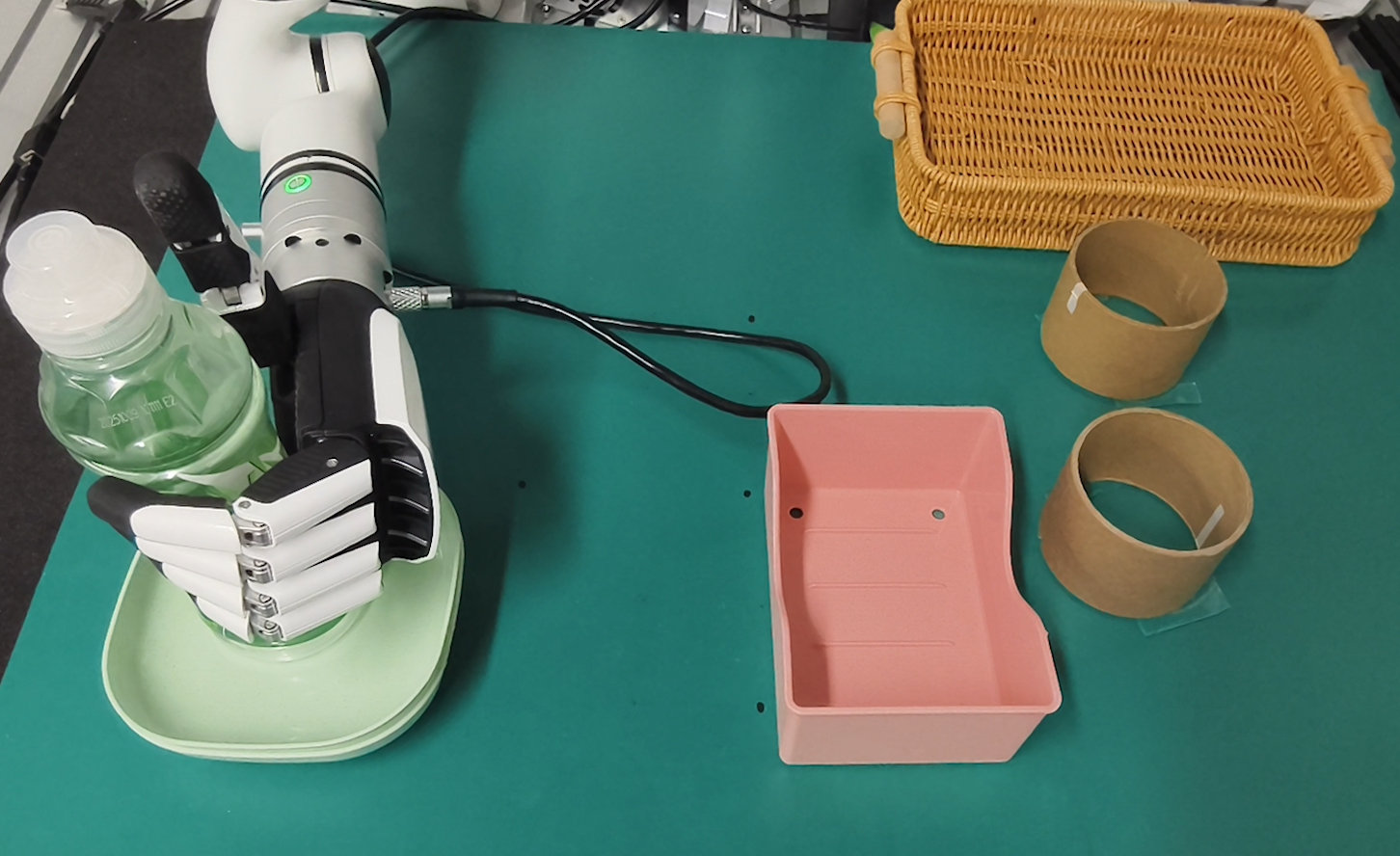} 
    \caption{\textbf{An example of the one-shot manipulation task.} The robot picks up the bottle from the table, and infers the category of the object based on both visual input and tactile information, and finally places the object to the right place accordingly.}
    \label{fig:appendix-task4}
\end{figure}

\subsection{Task Description}

\begin{figure}[t]
    \centering
    \includegraphics[width=0.48\textwidth]{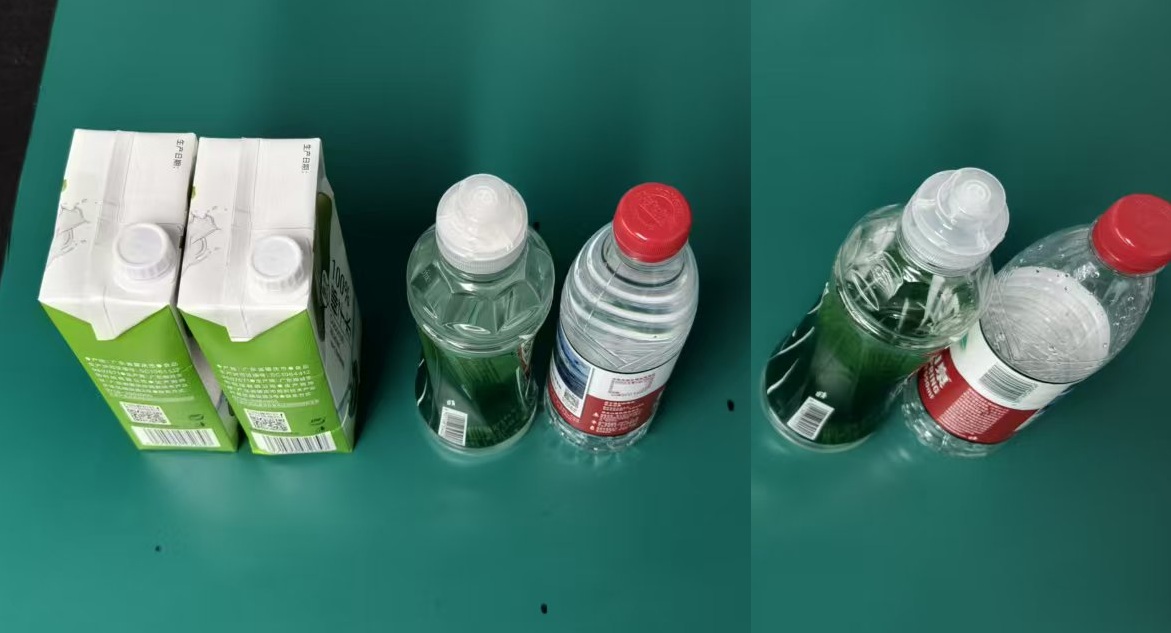} 
    \caption{\textbf{Six objects used in our one-shot manipulation task.} These objects are divided into three pairs of bottles, including both full and empty ones.}
    \label{fig:six-object}
\end{figure}

In our experiments~\cref{sec:experiments}, we choose five different tasks varying from perceptions on the numerical magnitude of tactile values to perceptions on the morphological distributions of tactility, among which four are manipulation tasks:

\begin{enumerate}
    \item \textbf{SoftHardPickPlace.} In this task, we divide our objects into two categories according to hardness. We choose two soft objects and two hard ones as seen objects used to collect human demonstrations, with another two soft objects and two hard ones as unseen objects. We prepare two baskets, one on the left and the other on the right. The robot (arm and hand) picks the object up with only tactility and proprioception, and places the object into a basket. This task is considered success only if the object is placed into the right basket (soft in the left, hard in the right). A demonstration example is shown in~\cref{fig:appendix-task1}. 
    \item \textbf{ObjectLocating.} Such a task requires the robot to find the correct location of the object after a pre-grasping contact. For each test time, the end-effector moves to a fixed location and touches the object on the table. The object is initially placed within a rectangular range of 10 cm $\times$ 20 cm, so that for each test, objects at different locations will result in different regions of tactile activation on the dexterous hand. For instance, in some cases the tactile regions on the index finger are activated, while in other cases the object is touched by the hand on the pinky. Once touching, the policy makes decisions representing the movement of the end-effector according to tactile information in the inference phase. The aim is to move the end-effector towards the correct direction with a proper distance so that the object can be grasped successfully with a direct grasp after moving the end-effector. A demonstration example is shown in~\cref{fig:appendix-task2}.

\begin{table*}[t]
  \centering
  \caption{\textbf{Basic settings of each manipulation task.} We list the inputs and outputs of the policies in our manipulation tasks: \texttt{SoftHardPickPlace}, \texttt{ObjectLocating}, \texttt{CompliantControl}, and the last one-shot manipulation task.}
  \label{tab:IO}
  \resizebox{0.99\linewidth}{!}{
      \begin{tabular}{ccccc}
        \toprule
        \textbf{Task} & \textbf{Input State} & \textbf{Output Action} \\
        \midrule
        SoftHardPickPlace & Tactility + Proprioception (3-DoF End-Effector Location) & 3-DoF End-Effector Location $(X, Y, Z)$ \\
        ObjectLocating & Tactility & 2-DoF End-Effector Location $(X, Y)$ \\
        CompliantControl & Tactility & 2-DoF End-Effector Delta Movement $(\Delta X, \Delta Y)$ \\
        One-Shot Manipulation & RGB Image + Tactility + Proprioception (arm joints) & 6-DoF arm joints \\
        \bottomrule
      \end{tabular}
  }
\end{table*}
    
    \item \textbf{CompliantControl.} In this task, the pose of the dexterous hand is fixed similar to a ``fist''. A person ``drags'' the hand towards a specific direction. Due to the clenching gesture of the ``fist'', dragging towards different directions will result in different regions of tactility on the hand to be activated. The policy infers the moving direction according to tactile information, which should be consistent with the intention of the person who applies external forces on the dexterous hand. A demonstration example is shown in~\cref{fig:appendix-task3}.
    \item \textbf{One-Shot Manipulation.} Such a task is used to test the capability of the policy when one-shot real robot demonstration is mixed with human data. Besides, we use both visual and tactile information, to verify the competence of tactility in terms of those cases where vision only is not enough. In this task, we choose three pairs of objects (six in total, as shown in~\cref{fig:six-object}). Within each pair, the two objects are similar in vision but have different physical properties. For instance, two identical bottles, one empty and another full of water, look similar but can result in different tactility due to their significant difference in weight. During test time, the robot picks up the object (\eg, a bottle) at a fixed location, and decides the following actions according to both the visual observation (from a third-person camera) and the tactile sense on the hand. The robot is required to place the object to the correct pre-defined location (each bottle full of water corresponding to a unique place; all empty bottles corresponding to an empty basket). A demonstration example is shown in~\cref{fig:appendix-task4}.
\end{enumerate}

For our manipulation tasks, \texttt{SoftHardPickPlace}, \texttt{ObjectLocating}, \texttt{CompliantControl}, as well as the last one-shot manipulation task, the inputs (states) and outputs (actions) of our policies are listed in~\cref{tab:IO}. For each task, we choose a specific suitable state representation and the corresponding action control.

\begin{figure}[t]
    \centering
    \includegraphics[width=0.3\textwidth]{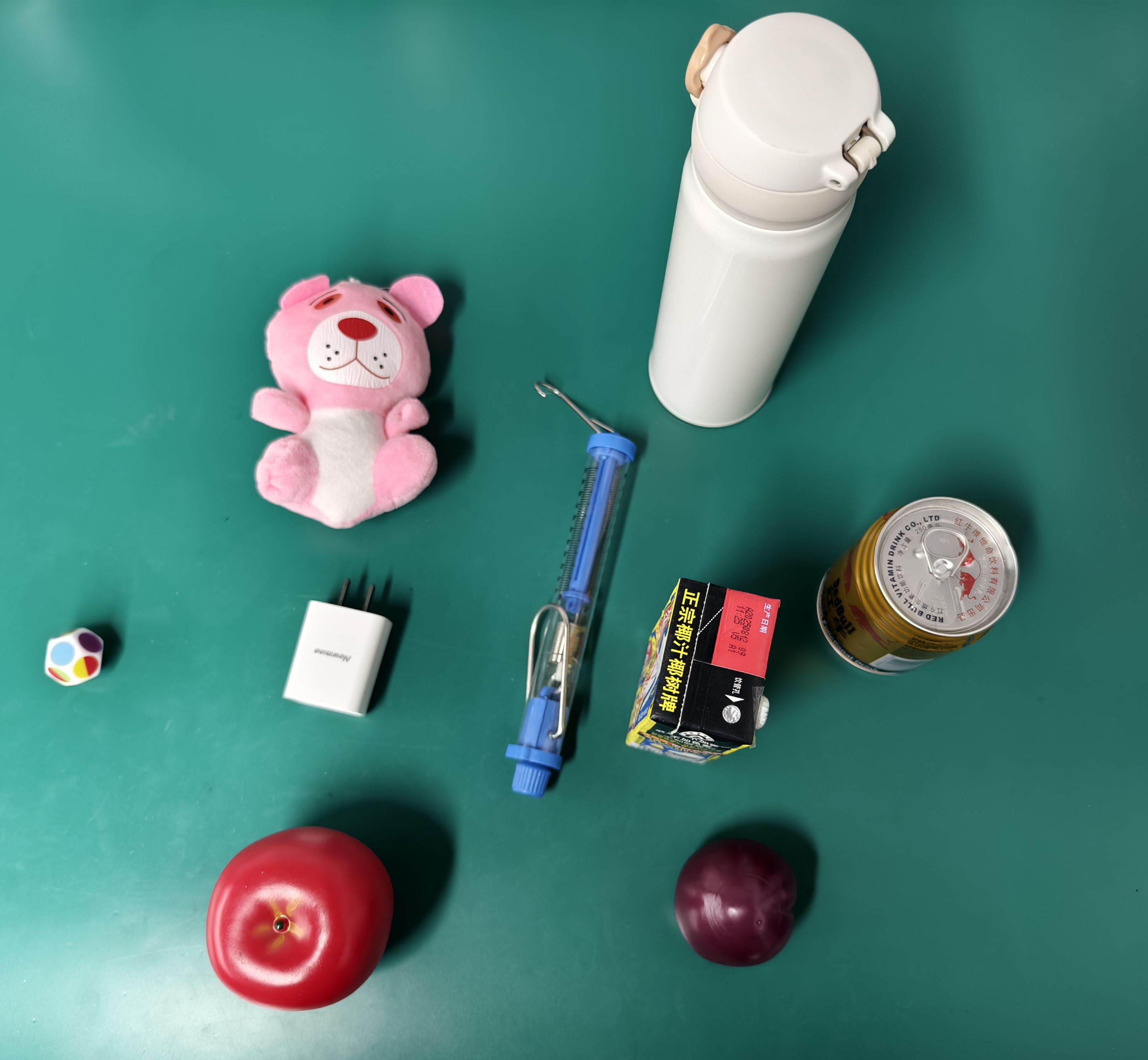} 
    \caption{\textbf{Ten categories (nine objects and another category identified as ``empty'') used in our \texttt{ObjectClassification} task.} These objects are unseen in the paired data used to train unified representations, ranging in various shapes, geometries, and physical features.}
    \label{fig:ten-object}
\end{figure}

Besides, we conduct experiments on object classification based on tactility over objects of different shapes, geometries, and physical features, denoted as our \texttt{ObjectClassification} task. In this task, we choose ten categories for classification (nine objects with another category identified as ``empty''), as listed in~\cref{fig:ten-object}.

\subsection{Training Policies for Tasks}

Here we present the details of training policies for each task in the expertiments~(\cref{sec:experiments}).

We list the implementation details of the four policies (networks and architectures) in~\cref{tab:implementation}. For our one-shot manipulation task, we utilize a pre-trained ResNet-18~\citep{resnet} as the visual backbone, and concatenate the visual embedding with the tactile embedding, which is then fed into the policy network.

\begin{table*}[t]
  \centering
  \caption{\textbf{Implementation details of the four policies.} We list the details including network architectures and hyperparameters of the policies in our manipulation tasks. $D_{\mathrm{tac}}$ is the dimension of tactility. For the following modalities: (1) raw data of tactile gloves, (2) raw data of Inspire Hand tactility, (3) our UV Map universal representation, (4) the embedding space of our unified representation learned from paired data, (5) the PatchMatch baseline that matches Inspire to gloves patch by patch, the dimensions are $D_{\mathrm{tac}}=137$, $D_{\mathrm{tac}}=1062$, $D_{\mathrm{tac}}=391$, $D_{\mathrm{tac}}=32$, $D_{\mathrm{tac}}=137$, respectively. }
  \label{tab:implementation}
  \resizebox{0.99\linewidth}{!}{
      \begin{tabular}{ccccc}
        \toprule
        \textbf{Implementation} & \textbf{SoftHardPickPlace} & \textbf{ObjectLocating} & \textbf{CompliantControl} & \textbf{One-Shot Manipulation} \\
        \midrule
        Network Architecture & MLP w/ Gumbel Routing & Multi-Head MLP & \makecell[c]{Tactile Encoder (MLP) \\+ Policy (MLP)} & \makecell[c]{Image Encoder (ResNet-18),\\ Tactile Encoder (MLP),\\ Policy (MLP)} \\
        \midrule
        Input Dim & $D_{\mathrm{tac}} + 3$ & $D_{\mathrm{tac}}$ & $D_{\mathrm{tac}}$ & $D_{\mathrm{tac}} + 6~(+~\mathrm{image})$ \\
        Hidden Dim & 512 & [512, 256] & 256 & 64 \\
        Output Dim & 3 & 2 & 3 & 6 \\
        Action Chunk Size & 16 & 64 & 1 & 16 \\
        Batch Size & 20 & 32 & 32 & 120 \\
        Learning Rate & 0.001 & 0.001 & 0.001 & $1\times 10^{-5}$ \\
        \bottomrule
      \end{tabular}
  }
\end{table*}

\section{Latent Space Analysis}
\label{sec:latent_analysis}

\begin{figure}[t]
    \centering
    \includegraphics[width=0.55\textwidth]{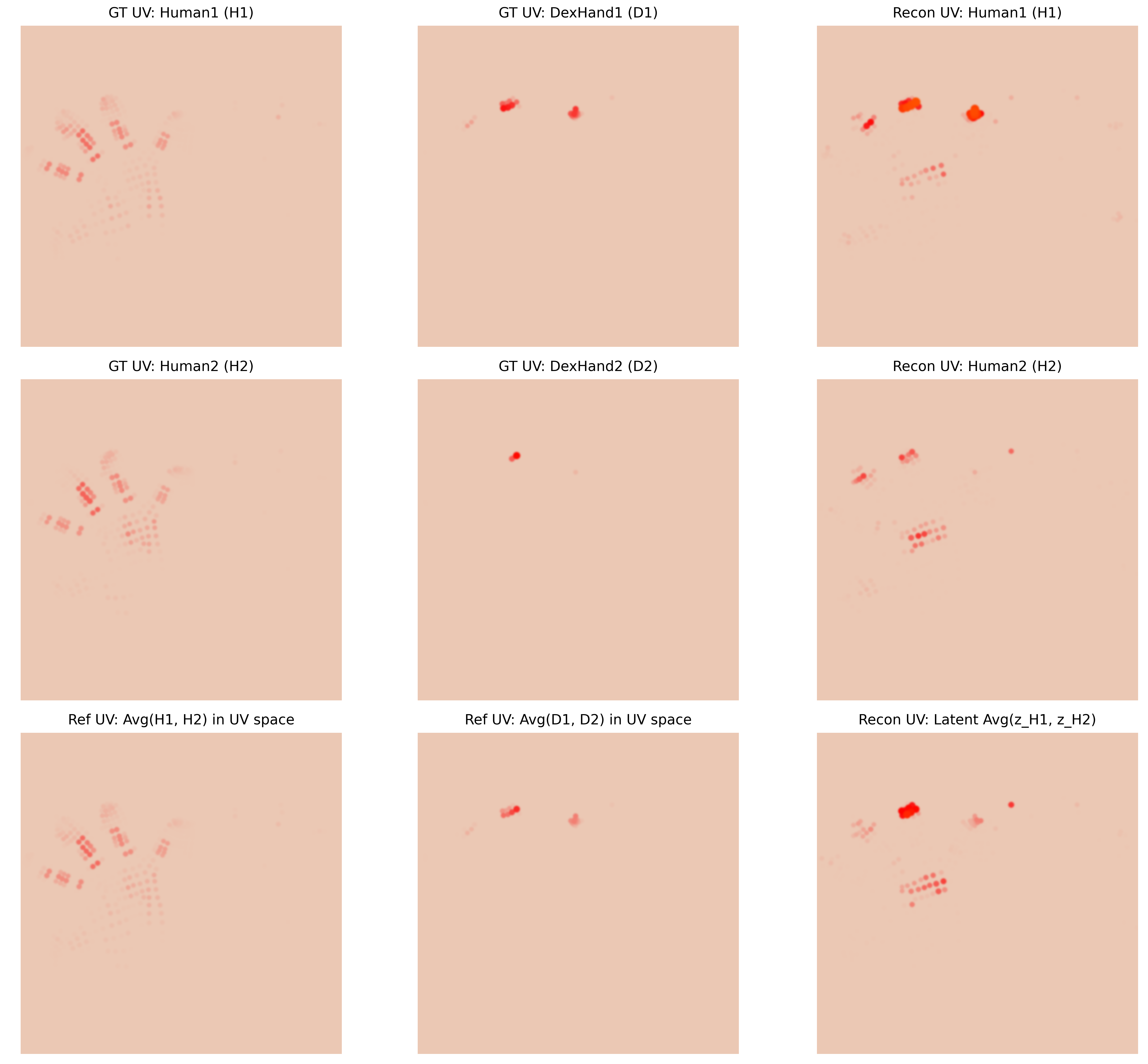} 
    \caption{\textbf{Visualization of latent space properties (sample A).} 
    This $3 \times 3$ grid demonstrates both reconstruction and linearity. 
    \textbf{(Right Col, Top \& Middle)}: The model accurately reconstructs robot tactile maps from human input, matching the ground truth in the Middle Column. 
    \textbf{(Right Col, Bottom)}: The decoded result of the averaged latent vector matches the physical average of the robot maps (Middle Col, Bottom), confirming the linearity and additivity of the latent space.}
    \label{fig:latent_vis1}
\end{figure}

\begin{figure}[t]
    \centering
    \includegraphics[width=0.55\textwidth]{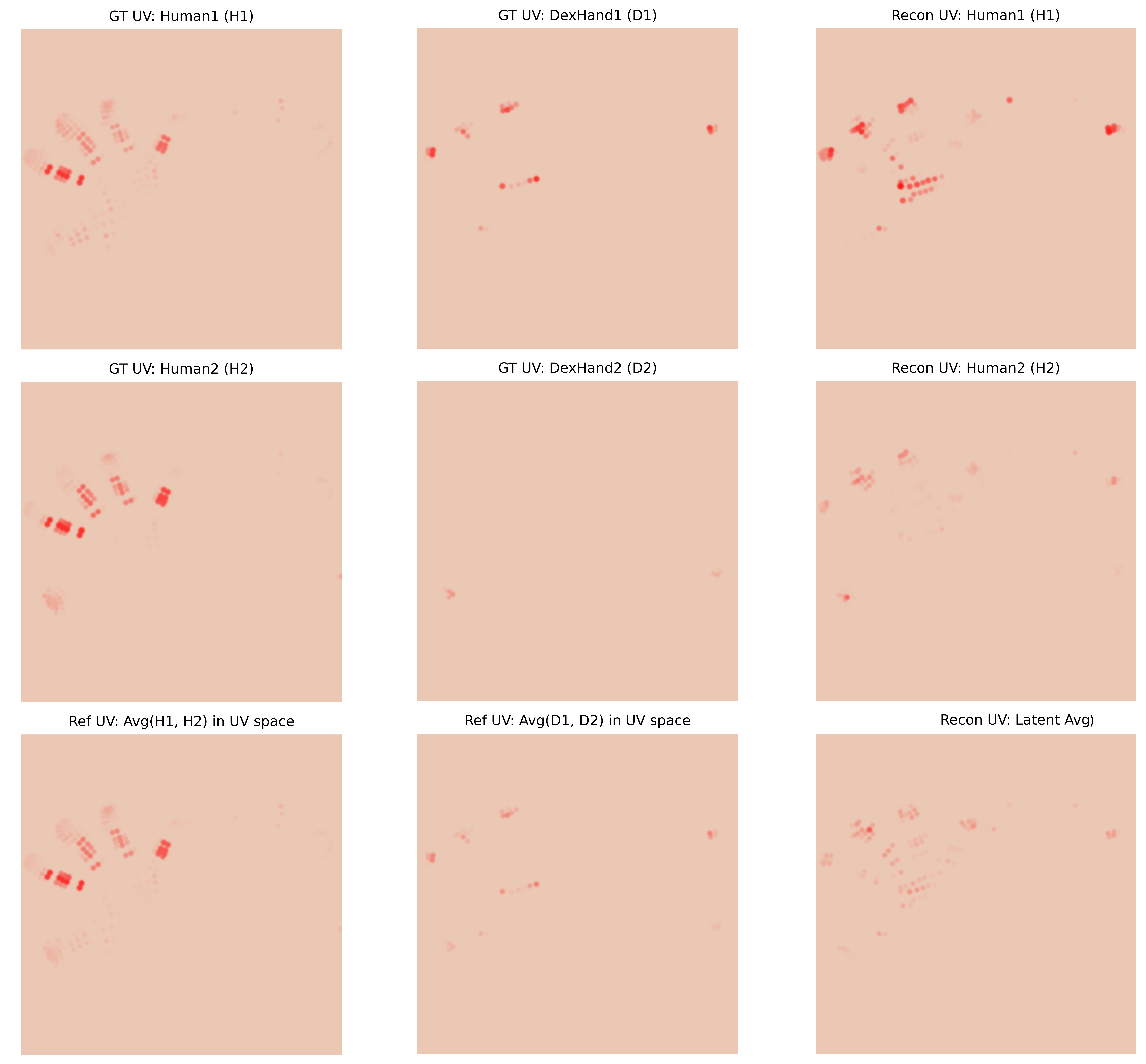} 
    \caption{\textbf{Visualization of latent space properties (sample B).} 
    Another example from the validation set. The consistent similarity between the decoded latent average (Bottom Right) and the ground truth average (Bottom Middle) reinforces that the learned representation captures the underlying continuous structure of tactile manipulation skills.}
    \label{fig:latent_vis2}
\end{figure}

To demonstrate the robustness and interpretability of the unified representation learned by UniTacHand, we visualize the latent space properties using paired human-robot data from the unseen validation set. We organize our analysis into two key aspects: cross-modal reconstructability and the linearity of the latent manifold.

Figures \ref{fig:latent_vis1} and \ref{fig:latent_vis2} present these properties using a $3 \times 3$ grid layout for two different pairs of object interactions. The layout is defined as follows:
\begin{itemize}
    \item \textbf{Left Column (Human Input):} Shows two distinct human tactile patterns ($H_1, H_2$) and their pixel-wise average ($(H_1+H_2)/2$).
    \item \textbf{Middle Column (Robot Ground Truth):} Shows the corresponding paired robotic tactile maps ($R_1, R_2$) and their pixel-wise average ($(R_1+R_2)/2$).
    \item \textbf{Right Column (Model Reconstruction):} The first two rows show the robot UV maps reconstructed purely from the human inputs (i.e., $D_R(E_H(H_i))$). The third row shows the decoding result of the \textit{averaged latent feature} (i.e., $D_R(\frac{z_1 + z_2}{2})$).
\end{itemize}

\subsection{Cross-Modal Reconstructability}
A critical requirement for human-to-robot transfer is the ability to map human tactile sensations to their robotic counterparts accurately. As seen in the first two rows of \cref{fig:latent_vis1} and \cref{fig:latent_vis2}, our model successfully performs this cross-modal translation.
The reconstructed robot maps (Right Column, Rows 1-2) closely resemble the ground-truth robot maps (Middle Column, Rows 1-2) in terms of contact location and intensity distribution. This indicates that the Human Encoder ($E_H$) effectively extracts a canonical representation that the Robot Decoder ($D_R$) can interpret, successfully bridging the morphological gap.

\subsection{Linearity and Additivity in the Latent Space}
Beyond simple reconstruction, a well-structured latent space should exhibit linearity, where algebraic operations in the latent space correspond to meaningful morphological transformations in the output space.
We investigate this by averaging the latent vectors of two distinct grasps: $z_{avg} = \frac{1}{2}(E_H(H_1) + E_H(H_2))$. We then decode this averaged vector to produce $\hat{R}_{avg} = D_R(z_{avg})$.

As illustrated in the third row of the figures, the decoded result from the latent average (Bottom Right) is morphologically consistent with the pixel-wise average of the ground truth robot data (Bottom Middle). This ``Additivity'' property suggests that our contrastive learning objective has shaped the latent manifold to be continuous and linear. Such a property is vital for generalization, as it implies the policy can smoothly interpolate between learned tactile primitives rather than overfitting to discrete training samples.





\clearpage

\end{document}